\documentclass[10pt,twocolumn,letterpaper]{article}

\usepackage[pagenumbers]{wacv} 

\usepackage{graphicx}
\usepackage{amsmath}
\usepackage{amssymb}
\usepackage{booktabs}
\usepackage{amsfonts} 
\usepackage{tabularx}
\usepackage{wrapfig}
\usepackage{caption}
\usepackage{adjustbox}
\usepackage{array}
\usepackage{amsthm}
\usepackage{graphics}
\usepackage{graphicx}
\usepackage{enumitem}
\usepackage{multirow}
\usepackage{wasysym}
\usepackage{soul}
\usepackage{svg}
\usepackage{tikz}
\usepackage{wasysym}
\usepackage{xspace}
\usepackage{siunitx}
\usepackage{mathtools}
\usepackage{algorithm}
\usepackage{algorithmicx}
\usepackage[noend]{algpseudocode}
\usepackage{subcaption}
\usepackage{xparse}
\usepackage{dsfont}
\usepackage{colortbl}
\usepackage{float}
\usepackage{bm}
\usepackage{pifont}
\usepackage{dashbox}
\usepackage{sourcecodepro}
\usepackage{longtable}
\usepackage{tcolorbox}

\let\olduparrow\uparrow
\renewcommand{\uparrow}[1][1pt]{%
  \mathrel{\raisebox{#1}{$\olduparrow$}}%
}

\newcommand{\ours}{\texttt{FAIR}\xspace}

\newcommand*{\code}[1]{{\fontfamily{sourcecodepro}\selectfont\textsl{#1}}}

\algnewcommand\algorithmicforeach{\textbf{for each}}
\algdef{S}[FOR]{ForEach}[1]{\algorithmicforeach\ #1\ \algorithmicdo}

\usepackage{graphicx}
\usepackage{amsmath}
\usepackage{amssymb}
\usepackage{booktabs}

\usepackage[pagebackref,breaklinks,colorlinks]{hyperref}

\usepackage[capitalize]{cleveref}
\crefname{section}{Sec.}{Secs.}
\Crefname{section}{Section}{Sections}
\Crefname{table}{Table}{Tables}
\crefname{table}{Tab.}{Tabs.}

\def\wacvPaperID{527} 
\def\confName{WACV}
\def\confYear{2026}

\begin{document}

\title{Towards Fine-Grained Adaptation of CLIP via a Self-Trained Alignment Score}

\author{
Eman Ali$^{1,3}$ \qquad
Sathira Silva$^{1}$ \qquad
Chetan Arora$^{2}$ \qquad
Muhammad Haris Khan$^{1}$ \\
$^{1}$Mohamed Bin Zayed University of Artificial Intelligence \qquad
$^{2}$IIT Delhi \qquad
$^{3}$Alexandria University
}

\maketitle

\begin{abstract}
Vision-language models (VLMs) like CLIP excel in zero-shot learning by aligning image and text representations through contrastive pretraining. 
Existing approaches to unsupervised adaptation (UA) for fine-grained classification with VLMs either rely on fixed alignment scores that cannot capture evolving, subtle class distinctions or use computationally expensive pseudo-labeling strategies that limit scalability.
In contrast, we show that modeling fine-grained cross-modal interactions during adaptation produces more accurate, class-discriminative pseudo-labels and substantially improves performance over state-of-the-art (SOTA) methods.
We introduce Fine-grained Alignment and Interaction Refinement (\ours), an innovative approach that dynamically aligns localized image features with descriptive language embeddings through a set of Class Description Anchors (CDA). This enables the definition of a Learned Alignment Score (LAS), which incorporates CDA as an adaptive classifier, facilitating cross-modal interactions to improve self-training in unsupervised adaptation.
Furthermore, we propose a self-training weighting mechanism designed to refine pseudo-labels in the presence of inter-class ambiguities.
Our approach, \ours, delivers a substantial performance boost in fine-grained unsupervised adaptation, achieving a notable overall gain of 2.78\% across 13 fine-grained datasets compared to SOTA methods.~\footnote{Our code will be made public upon acceptance.}.
\end{abstract}    
\section{Introduction}
\label{sec:intro}

\begin{figure}[!t]
    \centering
    \includegraphics[width=0.8\linewidth]{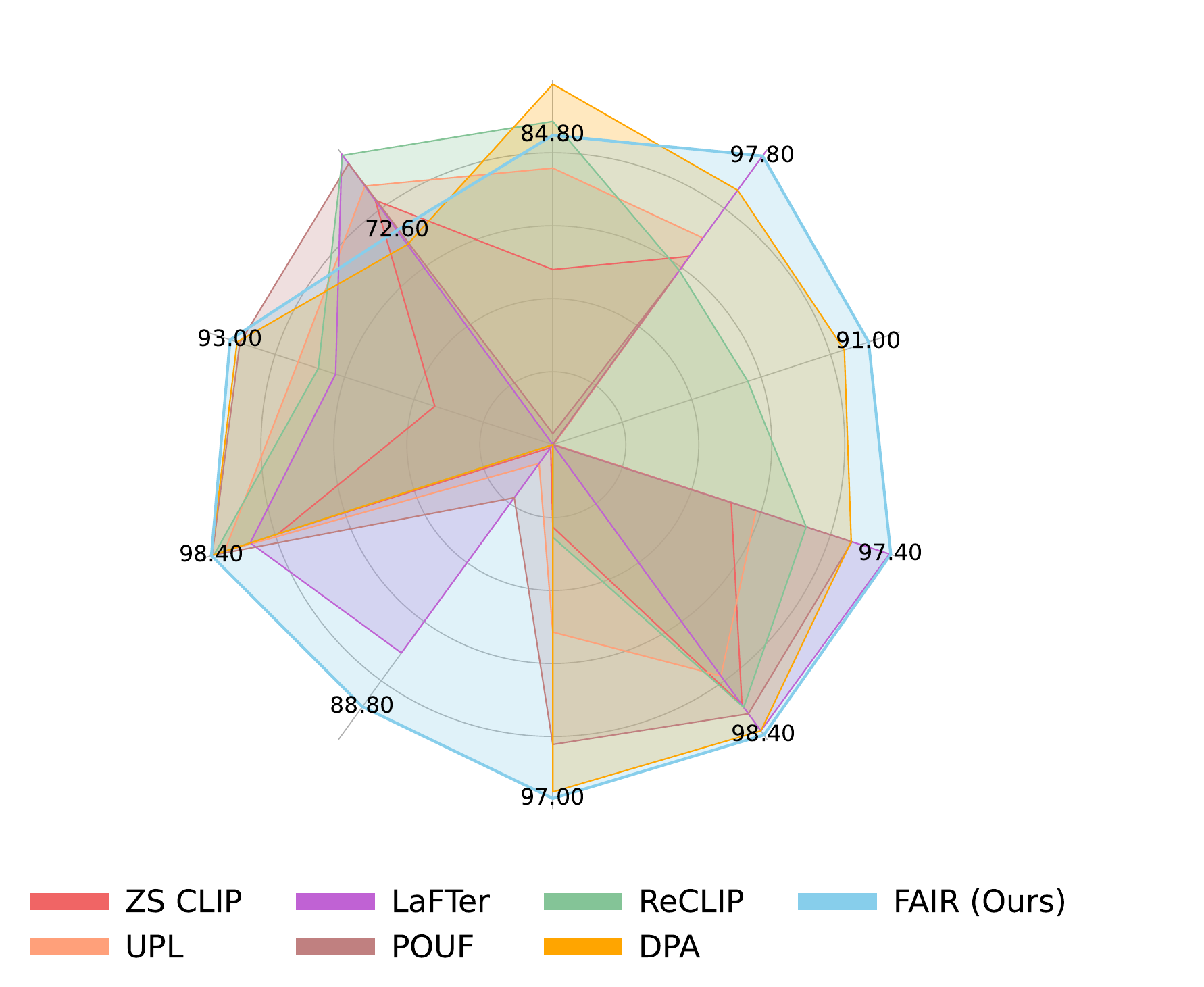}
    \caption{
    Class-wise classification accuracy (\%) comparison between \ours and state-of-the-art methods on the EuroSAT~\cite{helber2019eurosat} dataset. \ours achieves significant improvements, especially in visually and semantically similar classes where state-of-the-art methods struggle. \ours distinguishes fine-grained classes, demonstrating robust adaptation to challenging class distributions.    
    }
    \label{fig:radar_teaser}
\end{figure}

Large pretrained vision-language models (VLMs), such as CLIP~\cite{clip}, have demonstrated remarkable zero-shot capabilities by learning joint visual and textual representations through contrastive training on large-scale image–text pairs. This breakthrough has led to CLIP's widespread application across multiple computer vision tasks, including image classification~\cite{clip, khattakMaPLe, zhang2023prompt}, object detection~\cite{Gu2021OpenvocabularyOD, vidit2023clip, wu2023cora}, and semantic segmentation~\cite{Li2022LanguagedrivenSS, Xu2022GroupViTSS, lin2023clip}, all without task-specific supervision. 
However, adapting CLIP to fine-grained classification tasks, where visually similar classes must be distinguished, remains a significant challenge. In such tasks, CLIP's reliance on global image features and generic prompts (e.g., ``\code{a photo of a \{CLASS\}},'') often fails to capture the subtle visual cues necessary to separate fine-grained categories~\cite{yuksekgonul2023when, parcalabescu2021valse}. This leads to degraded performance on datasets such as EuroSAT, where different classes may share highly overlapping visual characteristics (see~\cref{fig:radar_teaser}: ZS CLIP).
At the same time, acquiring labeled training data for these domains can be costly or infeasible~\cite{zhang2024mediclip, huang2024adapting}. However, state-of-the-art (SOTA) methods often assume access to a set of target-domain class names, even in the absence of labeled images~\cite{upl, lafter, reclip, DPA}.

\begin{figure}[!t]
    \centering
    \includegraphics[width=\linewidth]{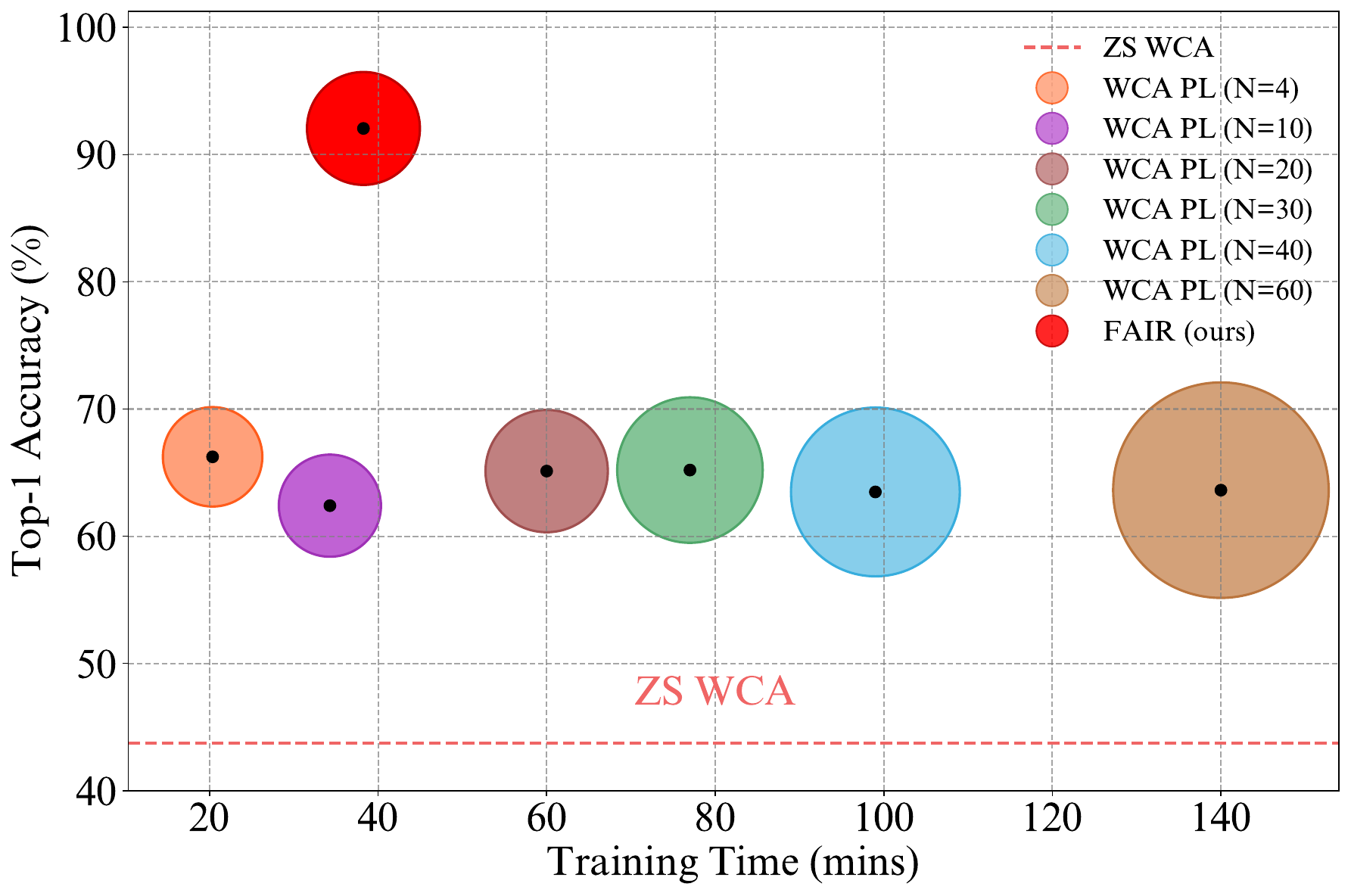}
    \caption{Efficiency comparison of \ours vs. WCA~\cite{wca} as a pseudo-labeling method using the EuroSAT dataset. The radius of each circle corresponds to the amount of GPU memory allocated.}
    \label{fig:wca_pl_crops}
\end{figure}

Recent works have proposed unsupervised adaptation (UA) of CLIP in this setting by utilizing Large Language Models (LLMs) to generate informative prompts or captions conditioned on class names~\cite{wca, CuPL, encoder}. For example, LaFTer~\cite{lafter} demonstrates that LLM-generated descriptions enable effective pseudo-labeling for adapting CLIP to fine-grained target datasets. LaFTer first uses these descriptions to fine-tune a classifier, which is then frozen and employed to generate pseudo-labels while fine-tuning CLIP's visual branch. 
However, a major limitation lies in its reliance on fine-tuned global image features, while the alignment scores are still computed using a fixed pretrained classifier.
In contrast, DPA~\cite{DPA} relies on fixed, handcrafted prompt ensembles, averaging them to create prototypical classifiers that are fine-tuned during adaptation. Similarly, ReCLIP~\cite{reclip} uses fixed prompts to refine the embedding space and label assignments through label propagation and cross-modality self-training. 
However, the dependence of both DPA and ReCLIP on manually crafted prompts limits their effectiveness, particularly on fine-grained datasets, as illustrated in~\cref{fig:radar_teaser}.
WCA~\cite{wca}, a recent zero-shot method, enhances the alignment score function by integrating LLM-generated captions with localized visual prompts (image crops), improving classification by aligning crop features with textual descriptions. However, its reliance on generating a large number of local image crops for pseudo-labeling significantly increases computational cost in unsupervised adaptation, as shown in~\cref{fig:wca_pl_crops}. These limitations motivate us to enhance WCA’s alignment score function to improve both efficiency and performance in CLIP’s unsupervised adaptation.

To this end, we introduce \textbf{F}ine-grained \textbf{A}lignment and \textbf{I}nteraction \textbf{R}efinement (\ours), a novel unsupervised adaptation framework for CLIP that introduces a Learned Alignment Score (LAS) based on selected local image crops and adaptive class representations. 
Our contributions can be summarized as follows:
\begin{itemize}    
    \item We propose a novel approach that combines selective crop relevance weighting and adaptive alignment scoring to enhance cross-modal alignment. Specifically, we (1) utilize the \texttt{[CLS]} token from the visual encoder to compute weights that rank image crops by relevance, selecting the top-$k$ most relevant crops for further alignment, and (2) introduce \textit{Class Description Anchors (CDA)} as an adaptive classifier. These components define \textit{Learned Alignment Score (LAS)}, enabling fine-grained interactions between the top-$k$ crops and CDA. This process, dubbed \textbf{F}ine-grained \textbf{A}lignment and \textbf{I}nteraction \textbf{R}efinement (\ours), significantly enhances pseudo-labeling accuracy and self-training.    
    \item We propose a novel weighting mechanism for the self-supervision loss function in \ours, to enhance pseudo-labeling accuracy in the presence of confusing classes. 
    \item Through extensive evaluation on 13 fine-grained datasets, \ours achieves a substantial gain of $2.78\%$, outperforming the state-of-the-art.   
\end{itemize}

\section{Related Work}
\label{sec:related_work}

\noindent\textbf{Vision-Language Models (VLMs):} 
Vision-Language Models (VLMs)~\cite{clip, ALBEF, blip, jia2021scaling, flava, florence, flamingo} demonstrate remarkable zero-shot performance in visual recognition tasks, owing to pretraining on vast image–text pairs from the web.
CLIP\cite{clip}, a pioneering VLM, excels in capturing vision–language correspondences, enabling zero-shot predictions across diverse inputs.
However, CLIP's performance is highly susceptible to prompting, making optimal prompt selection complex and time-consuming~\cite{coop}.
Recent research~\cite{coop, zhou2022learning, li2024promptkd, khattakMaPLe, lu2022prompt, cho2023distribution} has introduced methods to refine prompts using few-shot samples from target datasets, aiming to reduce reliance on manual prompt engineering. In contrast, our work proposes an entirely label-free approach to adapting CLIP to the target task.

\noindent\textbf{Unsupervised Adaptation for CLIP:} 
Despite advancements in supervised fine-tuning of CLIP, obtaining even a few labeled samples per class is often impractical in many applications~\cite{zhang2024mediclip, huang2024adapting}. To address this limitation, recent research has focused on fine-tuning CLIP with unlabeled samples from target datasets~\cite{upl, pouf}. However, these methods heavily rely on zero-shot CLIP's pseudo-labeling, which has shown limitations in fine-grained recognition tasks~\cite{yuksekgonul2023when, parcalabescu2021valse}.
CPL~\cite{zhang2024candidate} uses multiple candidate pseudo-labels as possible true labels instead of single hard labels. However, its effectiveness relies on candidate quality, as the true label may not be included among them.
LaFTer~\cite{lafter} leverages LLMs to generate text descriptions, aiding the model’s comprehension of nuanced visual–linguistic concepts.
However, LaFTer generates pseudo-labels by computing the alignment between learnable global image features and a frozen classifier pretrained on an LLM.
In this paper, we propose \ours, an innovative approach that enhances UA for CLIP by dynamically aligning localized image features with descriptive language embedding anchors through CDA.

\noindent\textbf{Leveraging LLMs for Enhanced CLIP Performance:} 
Large Language Models (LLMs)~\cite{floridi2020gpt, gallant1990perceptron, achiam2023gpt} are advanced language models with massive parameter sizes and exceptional learning capabilities\cite{qqqq}. One key feature of LLMs is in-context learning, where the model is trained to generate text based on a given context or prompt.
Recent approaches~\cite{lafter, encoder, wca, ProText, CuPL, fan2024improving, zhu2023not} have leveraged LLMs to enhance the generalization capabilities of CLIP by generating enriched text features for image categories.
For example, WCA~\cite{wca} utilizes LLM-generated class descriptions to create text features that align more effectively with image crop features rather than global image features.
In contrast to these approaches, our method leverages the semantic richness of text descriptions generated by LLMs to improve fine-grained classification.

\section{Methodology}
\label{sec:methodology}
\begin{figure*}[!t]
    \centering
    \includegraphics[width=\textwidth]{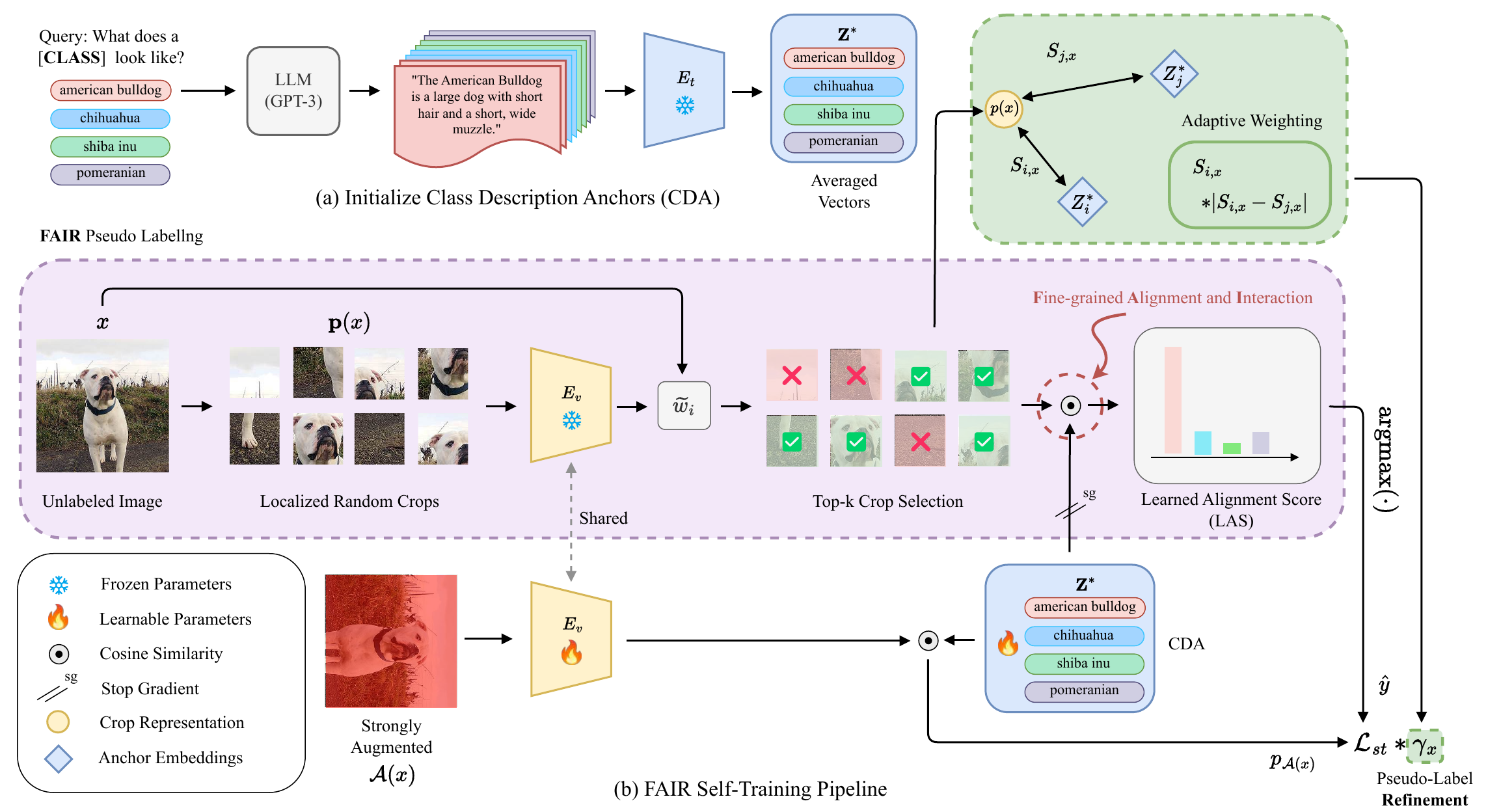}
    \captionsetup{singlelinecheck=off}
    \caption[foo bar]{Overall framework of \ours. 
    (a) shows the initialization of Class Description Anchors (CDA) using LLM-generated prompts, which are aggregated into representative features for each class. 
    (b) illustrates the \ours pseudo-labeling and self-supervision pipeline, where cross-modal fine-grained interaction and alignment (\textcolor[HTML]{B85450}{\dbox{circled}} in the figure)  between random image crops and CDAs generate pseudo-labels via a \textcolor[HTML]{82B366}{refined} alignment. 
    The pseudo-labeling branch shares parameters with the fine-tuned model.}
    \label{fig:main_architecture}
\end{figure*}

We address the problem of fine-grained image classification through UA of CLIP, using only unlabeled images from the target domain and a predefined set of class names associated with that domain.
Formally, given a target dataset \(\mathcal{D}_t=\left\{\mathcal{X}_t\right\}\), with unlabeled images \(\left\{x_i\right\}_{i=1}^U\), where \(x_i \in \mathcal{X}_t\), and a set of category names $y \in \mathcal{Y}$, our objective is to adapt a pretrained CLIP model, consisting of a visual encoder \(E_v\) and a textual encoder \(E_t\), to excel in the target task without relying on labeled examples.
To provide motivation and background context, we first review the alignment scores designed for zero-shot inference by CLIP, CuPL, and WCA in~\cref{subsec:zs_clip} through~\cref{subsec:zs_wca}. These functions determine how image features are matched with text embeddings, serving as the foundation for pseudo-labeling in UA. 
Building on these insights, we introduce \ours, a self-training framework that presents a learned alignment scoring function tailored for pseudo-label generation in fine-grained UA. We further enhance this framework with an adaptive weighting mechanism to handle ambiguous cases with visually similar classes. 
The overall architecture is illustrated in~\cref{fig:main_architecture}.

\subsection{Zero-shot Visual Classification in CLIP}
\label{subsec:zs_clip}

CLIP demonstrates remarkable zero-shot performance for image classification without requiring training on the target dataset. For inference on unlabeled images $x$ from the target dataset $\mathcal{D}_t$, CLIP employs its text encoder $E_t$ to generate embeddings for a set of potential class names $\mathcal{Y}$. Specifically, well-crafted prompts $\mathbf{t} = {t^y \mid y \in \mathcal{Y}}$ are input into $E_t$, producing natural language embeddings for each class, denoted as $\mathbf{Z} \in \mathbb{R}^{c \times d}$, where $c$ is the number of classes in $\mathcal{D}_t$ and $d$ is the feature space dimension. The zero-shot classification label $\hat{y}$ for a query image feature $f$ is obtained by first computing the CLIP alignment score, as defined in~\cref{eq:sim_func}, and then predicting the class using~\cref{eq:pseudo}.
\begin{gather}
    \psi_{\text{clip}}\left(x, t^y\right | E_v, E_t) =  \text{sim}(f, Z^y) \label{eq:sim_func}\\
    \hat{y} = \arg\max_{y \in \mathcal{Y}} (\psi_{\text{clip}}\left(x, t^y\right | E_v, E_t)) 
\label{eq:pseudo}
\end{gather}
where $\psi$ represents the cosine similarity function, and this notation will be used throughout the following sections to indicate a generalized alignment score function.

\subsection{Zero-shot Transfer in CuPL}
\label{subsec:zs_cupl}

CuPL~\cite{CuPL} replaces the handcrafted prompts $\textbf{t}$ in~\cref{eq:sim_func} with an ensemble of class-specific descriptions generated by an LLM, as shown in~\cref{eq:cupl_desc}. For each class label $y \in \mathcal{Y}$, the LLM $h(\cdot)$ generates multiple descriptions in response to the query ``\code{what does a \{$y$\} look like?}''(~\cref{eq:llm_cupl}). The CLIP similarity function is then redefined by averaging the alignment scores across these descriptions, as formulated in~\cref{eq:cupl_average}. 
\begin{gather}
    \textbf{t} = \{h(y)\}_{y \in \mathcal{Y}}, \quad \label{eq:cupl_desc} \\
    h(y) = \{t^y_j\}_{j=1}^{M}, \label{eq:llm_cupl} \\
    \psi_{\text{CuPL}}\left(x, \textbf{t}^y\right | E_v, E_t) = \frac{1}{M} \sum_{j=1}^M \text{sim}(f , \mathbf{Z}^y_j),
    \label{eq:cupl_average}
\end{gather}
where $M$ denotes the total number of generated descriptions per class, $\textbf{Z}^y \in \mathbb{R}^{M \times d}$ represents the description embeddings matrix for class $y$, where each row $\textbf{Z}^y_j = E_t(t^y_j)$ corresponds to the embedding of the $j^{th}$ description for class $y$.

\subsection{Zero-shot Transfer in WCA}
\label{subsec:zs_wca}

WCA~\cite{wca} proposes an alignment score function, where alignment between fine-grained textual descriptions and local areas within an image is considered. 
Given an image $x \in \mathbb{R}^{H \times W \times 3}$, the localized crops, denoted as $\mathbf{p}(x)$, are created as shown in~\cref{x_patch}.
\begin{equation}
    \label{x_patch}
    \mathbf{p}(x) = \left\{ p_i = \phi(x, \lambda_i \min(W, H)) \mid i = 1, \dots, N \right\}
\end{equation}
where $\phi(\cdot)$ performs a random crop on the input image, $\lambda_i$ is a random variable drawn from a uniform distribution $\lambda_i \sim U(\alpha, \beta)$, and $N$ denotes the number of crops generated per image. Then, WCA evaluates the similarities between localized image crops $\mathbf{p}(x)$ and text descriptions $h(y)$, yielding a matrix-valued function $\Theta(x,y|p,h,E_v,E_t)$ as follows:
\begin{equation}
\label{eq:wca_cross-alignment}
    \Theta =
    \begin{bmatrix}
        \text{sim}(f_1, \textbf{Z}^y_1) & \cdots & \text{sim}(f_1, \textbf{Z}^y_{M})\\
        \vdots & \ddots & \vdots\\
        \text{sim}(f_{N}, \textbf{Z}^y_1) & \cdots & \text{sim}(f_{N}, \textbf{Z}^y_{M})
    \end{bmatrix}
\end{equation}
where $f_i = E_v(p_i)$ are the local crop features.
Instead of a naive average of $\Theta$, WCA introduces image-to-image and text-to-text relevance-based weights. Specifically, weights for image crops, denoted as \(\mathcal{W} = \{w_i\}_{i=1}^{N}\), are computed by comparing local crop view features to the global view feature $f$. Similarly, for text descriptions, weights \(\mathcal{V} = \{v_j\}_{j=1}^{M}\) are computed by comparing text descriptions with the prompt template ``\code{a photo of a \{y\}}'' as follows:
\begin{gather}
    w_i = \frac{\exp\left(\text{sim}(f, f_i)\right)}{\sum_{l=1}^{N} \exp\left(\text{sim}(f, f_{l})\right)} \label{eq:w_i}\\
    v_j = \frac{\exp\left(\text{sim}(y, t^y_j |E_t)\right)}{\sum_{l=1}^{M} \exp\left(\text{sim}(y, t^y_l |E_t)\right)} \label{eq:v_j}
\end{gather} 
The alignment score function for WCA is then defined as:
\begin{equation}
    \psi_{\text{WCA}}\left(x, \textbf{t}^y\right | \mathbf{p}, E_v, E_t) = \sum_{i=1}^{N} \sum_{j=1}^{M} w_i v_j \Theta_{ij} \label{eq:wca_sim_func}
\end{equation} 
Finally, the zero-shot prediction is computed similarly to~\cref{eq:pseudo}.

\subsection{\ours for Unsupervised Adaptation of CLIP}
\label{subsec:fair_adaptation}

\begin{figure}[!t]
    \centering
    \includegraphics[width=\linewidth]{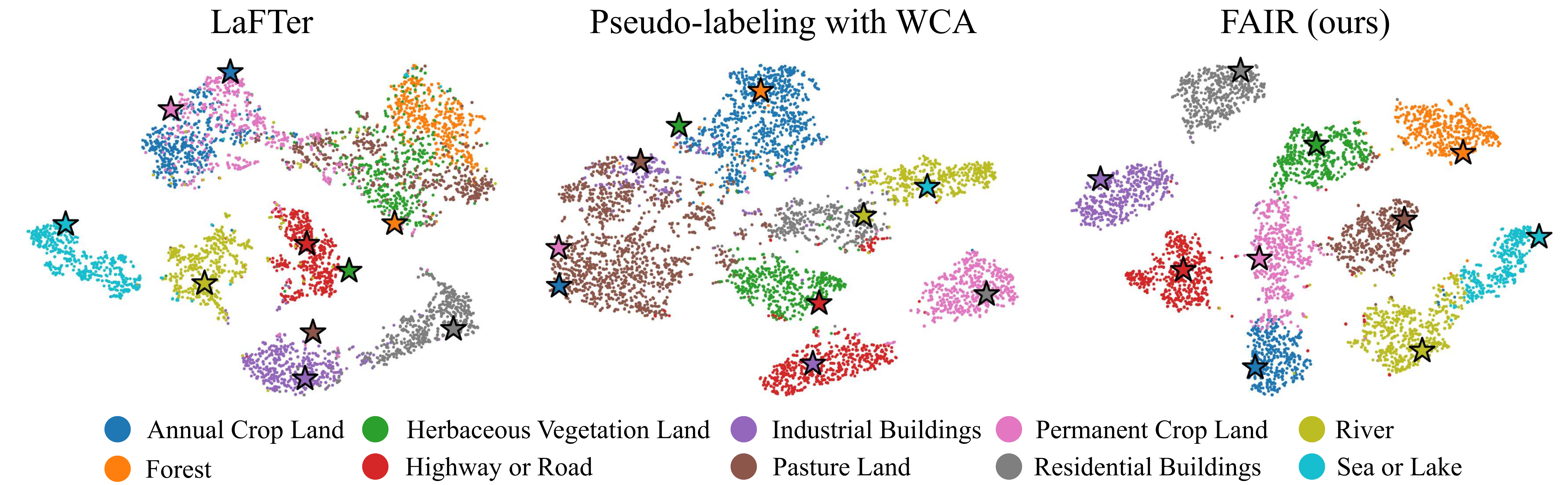}
    \caption{t-SNE plots of image embeddings (circle $\newmoon$) for LaFTer~\cite{lafter}, WCA~\cite{wca} as a pseudo-labeling method, and \ours, along with their corresponding classifiers (star $\bigstar$) on the EuroSAT dataset. LaFTer uses fixed alignment-based pseudo-labels, causing class confusion, while \ours leverages LAS to better separate class clusters and enhance discriminability.} 
    \label{fig:tsne}
\end{figure}

\begin{figure}[t!]
    \centering
    \begin{subfigure}[b]{0.22\textwidth} 
        \centering
        \includegraphics[width=\textwidth]{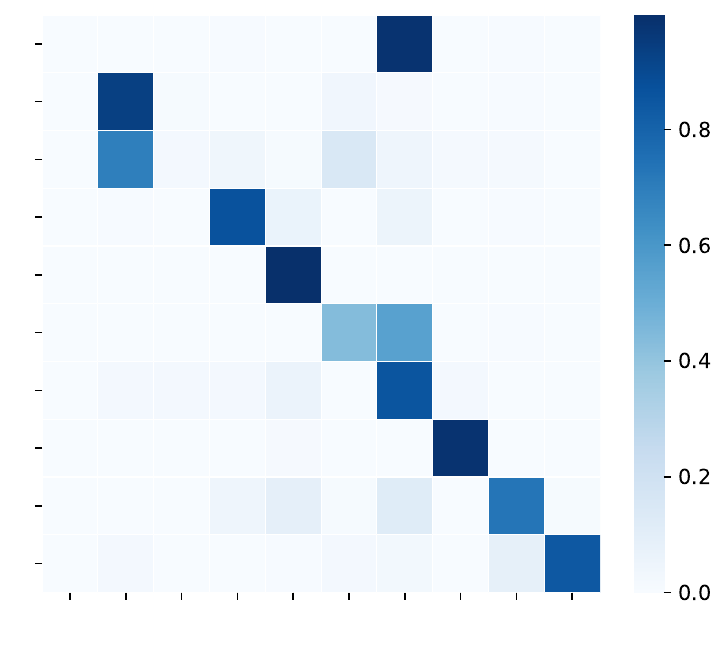}
        \caption{}
        \label{fig:conf_fix}
    \end{subfigure}
    \hspace{0.01\textwidth} 
    \begin{subfigure}[b]{0.22\textwidth}
        \centering
        \includegraphics[width=\textwidth]{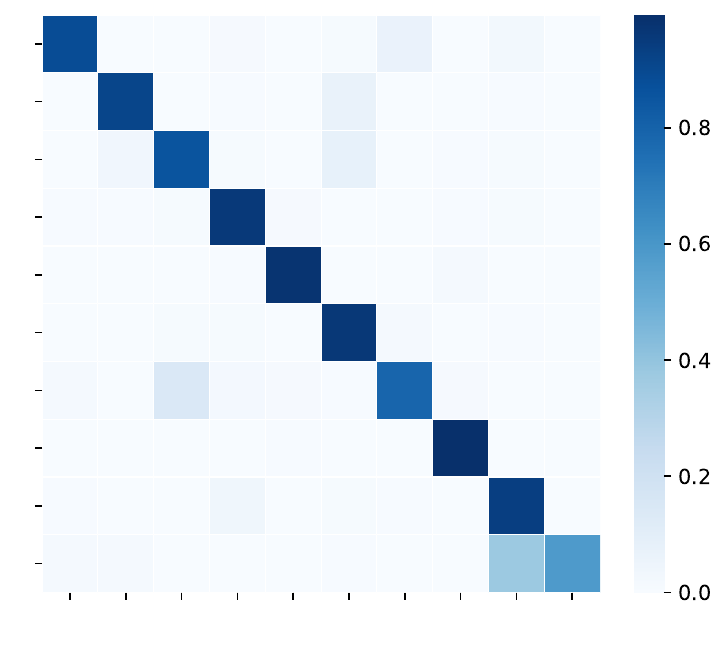}
        \caption{}
        \label{fig:conf_fair}
    \end{subfigure}
    \caption{Comparative confusion matrices illustrating model performance on the EuroSAT dataset: (a) using a fixed alignment score (LaFTer), and (b) using a learned alignment score (\ours).}   
    \label{fig:conf_eurosat}
\end{figure}

\textbf{Motivation:} While WCA~\cite{wca} makes a significant stride in zero-shot fine-grained classification through localized visual prompting and detailed text descriptions, it faces limitations when employed to generate pseudo-labels for UA tasks. One major challenge is the high computational demand of generating numerous image crops (approximately 60) to capture fine-grained details, which constrains its scalability within the UA framework. Moreover, WCA utilizes all randomly sampled crops for cross-alignment, which can reduce prediction accuracy, even with the use of image-to-image similarity $\mathcal{W}$ (\cref{fig:wca_pl_crops}), and necessitates a forward pass through the model for each crop.
On the other hand, LaFTer~\cite{lafter}, a UA method for CLIP, uses LLM-generated text descriptions to train a classifier, which is then frozen and used to generate pseudo-labels throughout training. However, this fixed alignment-based pseudo-labeling strategy limits LaFTer’s effectiveness on fine-grained tasks, as computing alignment scores between the classifier and global image features results in significant confusion between visually similar classes, as shown in~\cref{fig:conf_eurosat,fig:tsne}.
The key challenge in CLIP-based UA lies in aligning class-discriminative visual features with LLM-generated text representations. This highlights the need for a new approach that integrates WCA's localized visual prompting with the power of LLMs to better handle fine-grained unsupervised adaptation.

\noindent
\textbf{Pseudo-labeling via a Learned Alignment Score:} This observation motivated us to develop a more flexible and adaptive approach for UA tasks. Instead of relying on a fixed pretrained classifier, we introduce a learned alignment score that can be updated concurrently with the image encoder. Our aim is to maintain a dynamic equilibrium between the two modalities throughout the training process. 
This novel approach, which we refer to as \textbf{F}ine-grained \textbf{A}lignment and \textbf{I}nteraction \textbf{R}efinement (\ours), addresses the limitations of UA methods in fine-grained classification tasks and paves the way for more robust and accurate cross-modal alignment, especially between fine-grained features.
By allowing image and text representations to adapt during training, \ours captures more nuanced and evolving fine-grained semantic relationships between visual and textual modalities, leading to improved classification performance.~\cref{fig:main_architecture} illustrates the detailed self-training framework of \ours.

We identify two potential enhancements in WCA, making it more effective for adaptation in a self-training pipeline. First, the current crop weight calculation in~\cref{eq:w_i} utilizes image-to-image similarity based on CLIP's unified visual feature in the multi-modal space. Prior research~\cite{caron2021emerging} suggests that the \texttt{[CLS]} token more effectively captures comprehensive global semantic and structural details, making it often more suitable for image-based similarity tasks.
Second, the crop-description-based alignment score in~\cref{eq:wca_sim_func} could be refined further by filtering out irrelevant crops based on their weights, enhancing the focus on the most relevant visual information.

\begin{figure}[t!]
    \centering
    \includegraphics[width=0.35\textwidth]{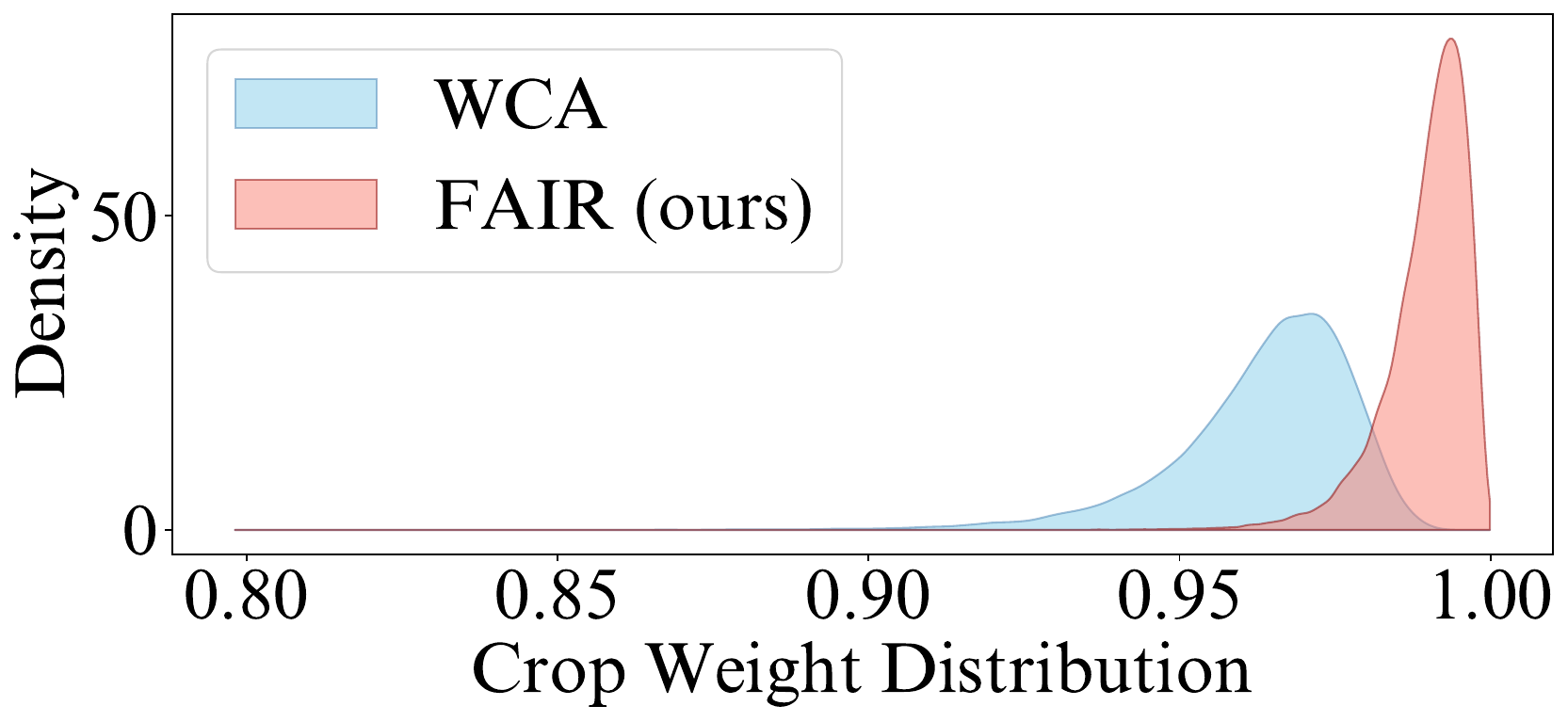}
    \caption{Overview of the density distribution of crop weights in \ours (ours), compared to WCA~\cite{wca}, over the Flowers dataset.}
    \label{fig:crop_weight_dist}
\end{figure}

\begin{figure}[ht!]
    \centering
    \begin{subfigure}[b]{0.225\textwidth} 
        \centering
        \includegraphics[width=\textwidth]{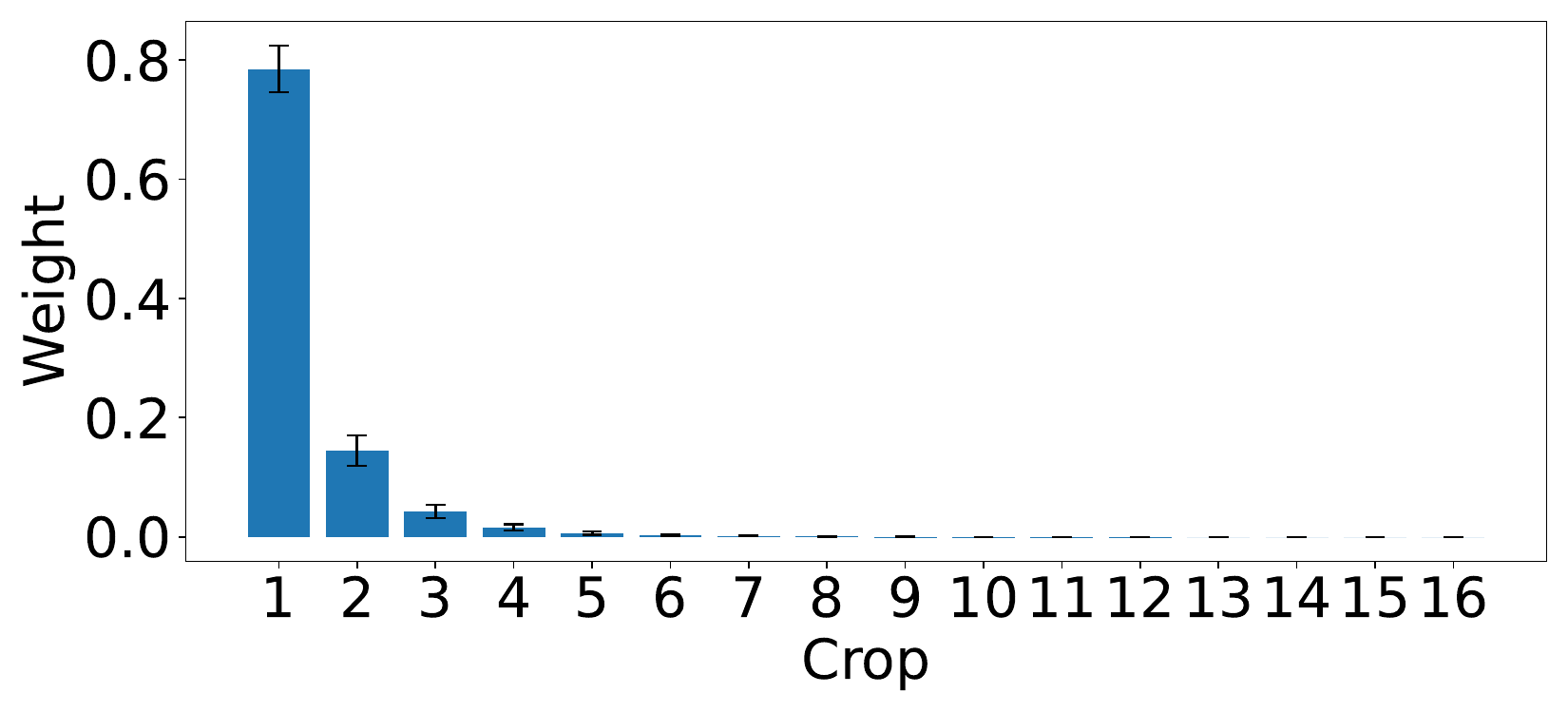}
        \label{fig:softmax_norm}
    \end{subfigure}
    \begin{subfigure}[b]{0.225\textwidth}
        \centering
        \includegraphics[width=\textwidth]{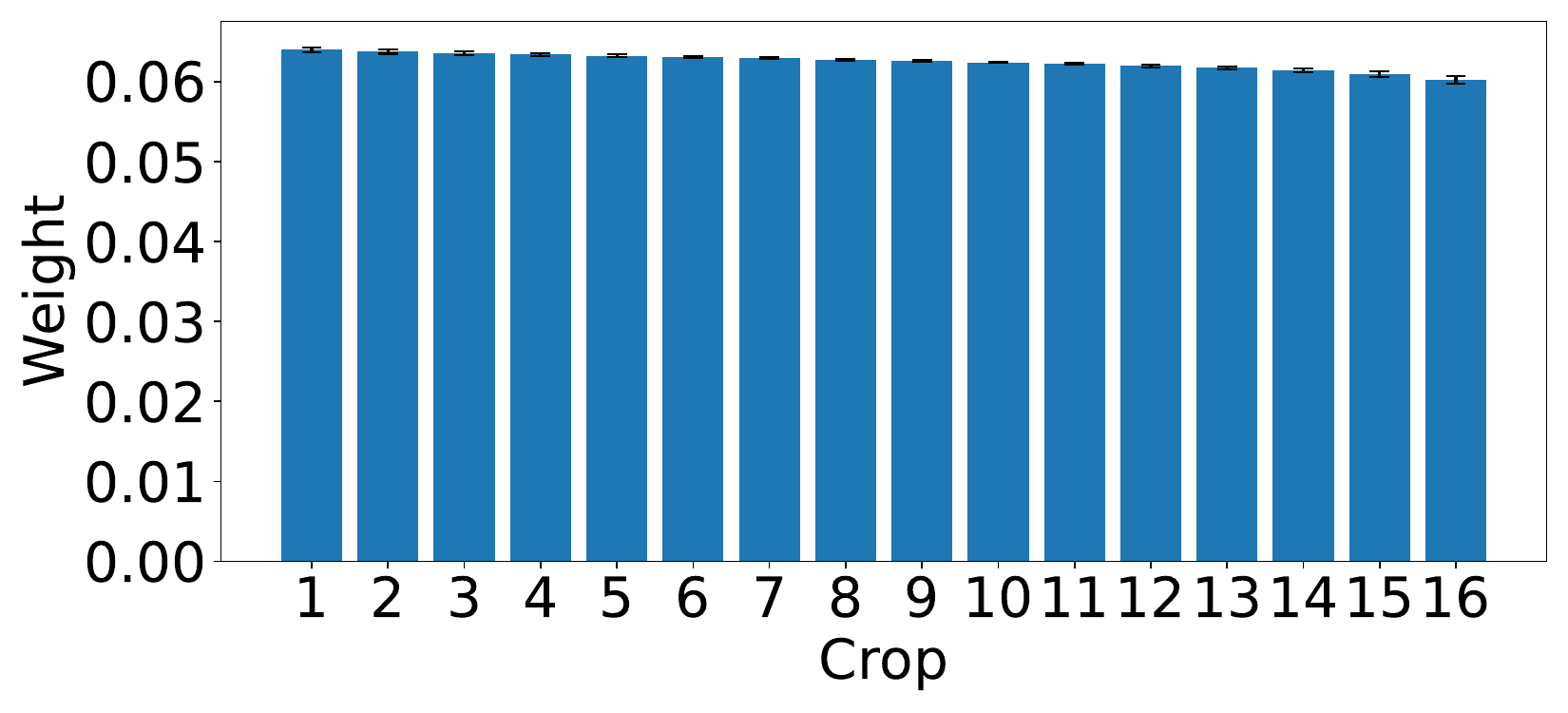}
        \label{fig:no_softmax_norm}
    \end{subfigure}
    \vspace{-7mm} 
    \caption{Comparison between the crop weights with (left) and without (right) softmax normalization.  Softmax is less effective when selecting top-$k$ crops from $N$.}
    \label{fig:crop_weights_justification} 
\end{figure} 

\noindent We empirically demonstrate that the aforementioned claims are validated in~\cref{fig:crop_weight_dist}, which compares the crop weight density distributions between WCA and \ours. The figure highlights the effectiveness of our approach in selecting high-quality crops, revealing that the crops chosen by \ours are more closely aligned with the specific global view. Based on these observations, we refine the crop weights defined in~\cref{eq:w_i}. As illustrated in~\cref{eq:w_i_theta}, we get rid of the softmax normalization since it can excessively emphasize high weights due to the exponential function, as shown in~\cref{fig:crop_weights_justification}. 

Moreover, we introduce Class Description Anchors (CDA), $\textbf{Z}^{*} \in \mathbb{R}^{c\times d}$, instead of using text descriptions $h(y)$, as in~\cref{eq:wca_cross-alignment}. These embeddings are initialized as an average of the text descriptions generated by the LLM, and the text encoder is discarded. 
Then, the similarities between localized image crops $\mathbf{p}(x)$ and CDA $\textbf{Z}^{*}$ yield a matrix-valued function $\widetilde{\Theta}(x,\textbf{Z}^{*}|p,E_v) \in \mathbb{R}^{N\times C}$, as expressed in~\cref{eq:wca_LTIA_learnable}.
\begin{equation}
\label{eq:wca_LTIA_learnable}
\widetilde{\Theta} =
\begin{bmatrix}
\text{sim}(f_1, \textbf{Z}^{*}_1) & \cdots & \text{sim}(f_1, \textbf{Z}^{*}_c) \\
\vdots & \ddots & \vdots \\
\text{sim}(f_{N}, \textbf{Z}^{*}_1) & \cdots & \text{sim}(f_{N}, \textbf{Z}^{*}_c)
\end{bmatrix}
\end{equation}
Accordingly, the relevance weight for the image crops, denoted as \(\mathcal{\widetilde{W}} = \{\widetilde{w_i}\}_{i=1}^{N}\), is given by:

\begin{table*}[t]
    \centering
    \scriptsize
    \setlength{\tabcolsep}{3pt} 
    \renewcommand{\arraystretch}{1.1} 
    \begin{tabularx}{\textwidth}{l >{\centering\arraybackslash}p{1cm} *{14}{>{\centering\arraybackslash}X}}
      \toprule
      \textbf{Method} & \textbf{Venue} & 
      \textbf{Caltech} & \textbf{DTD} & \textbf{EuroSAT} & \textbf{Food} &
      \textbf{Flowers} & \textbf{OxPets} & \textbf{StCars} & \textbf{UCF101} & 
      \textbf{CUB} & \textbf{RES45} & \textbf{CIFAR100} & \textbf{SUN} &      
      \textbf{AID} & \textbf{Avg} \\
      \midrule
      
      \rowcolor[gray]{0.9} \multicolumn{16}{c}{\textbf{Zero-shot Methods}} \\
      \midrule
      CLIP~\cite{clip} & ICML'21 & 92.60 & 44.63 & 48.90 & 83.95 & 66.42 & 87.46 & 58.70 & 61.86 & 51.76 & 57.59 & 64.47 & 61.32 & 64.70 & 64.95  \\
      CuPL~\cite{CuPL} & ICCV'23 & \underline{94.62} & 50.11 & 50.06 & 84.05 & 69.51 & 87.16 & 60.79 & \underline{66.90} & 49.71 & 61.14 & 65.22 & 65.57 & 59.80 & 66.51 \\ 
      WCA$^{*}$~\cite{wca} & ICML'24 & 93.97 & 52.02 & 43.44 & 84.35 & 68.90 & 86.56 & \textbf{62.99} & 66.67 & 51.76 & 62.78 & 52.77 & 65.23 & 59.33 & 65.44 \\  
      
      \midrule
      \rowcolor[gray]{0.9} \multicolumn{16}{c}{\textbf{UA Methods}} \\
      \midrule
      
      UPL~\cite{upl} & - & 92.36 & 45.37 & 51.88 & 84.25 & 67.40 & 83.84 & 49.41 & 62.04 & 49.22 & 57.63 & 67.41 & 62.12 & 68.27 & 64.71 \\
      POUF~\cite{pouf} & ICML'23 & 94.10 & 46.10 & 62.90 & 82.10 & 67.80 & 87.80 & 57.70 & 61.20 & 51.59 & 66.40 & 62.00 & 60.00 & 69.50 & 66.86 \\
      LaFTer~\cite{lafter} & NeurIPS'23 & 94.39 & 50.32 & 69.96 & 82.45 & 72.43 & 84.93 & 57.44 & 65.08 & 37.66 & 61.60 & 69.79 & 65.87 & 53.70 & 66.59 \\
      ReCLIP$^{\dagger}$~\cite{reclip} & WACV'24 & 93.84 & 53.88 & 69.48 & 84.22 & 72.63 & 87.11 & 58.84 & 66.67 & 53.95 & \underline{73.05} & 71.43 & 58.27 & 76.13 & 70.73 \\
      DPA~\cite{DPA} & WACV'25 & \textbf{95.94} & \underline{55.96} & \underline{79.94} & \underline{84.76} & \underline{75.56} & \underline{90.11} & 56.83 & 66.69 & \textbf{56.70} & 71.11 & \underline{74.22} & \textbf{68.13} & \underline{81.10} & \underline{73.62} \\
      
      \rowcolor[HTML]{FFD6E7} \textbf{\texttt{\ours}} (ours) & - & 94.12 & \textbf{62.07} & \textbf{91.92} & \textbf{84.89} & \textbf{75.72} & \textbf{90.52} & \underline{61.83} & \textbf{73.54} & \underline{55.25} & \textbf{73.27} & \textbf{76.17} & \underline{67.95} & \textbf{85.97} & \textbf{76.40} \\ 
      
      \bottomrule
    \end{tabularx}
    
  \caption{Top-1 accuracy (\%) comparison across 13 datasets for state-of-the-art methods using the ViT-B/32 backbone. 
  $*$ indicates results reproduced using the same number of crops as \ours. 
  ReCLIP\textsuperscript{\dag} denotes results obtained by training ReCLIP under an inductive setting.
  }
  \label{table:sota_vit_b_32}
\end{table*}

\begin{gather}
    \widetilde{w_i} = \frac{\text{sim}(f^{\texttt{[CLS]}}, f_i^{\texttt{[CLS]}})}{\sum_{l=1}^{N}\text{sim}(f^{\texttt{[CLS]}}, f_{l}^{\texttt{[CLS]}})} \label{eq:w_i_theta} \\
    \mathcal{I}_k = \text{argsort}(\mathcal{\widetilde{W}})[:k] \label{eq:crop_selection}
\end{gather}
\Cref{eq:crop_selection} describes the crop selection based on \cref{eq:w_i_theta}, where $k$ represents the top-$k$ crops selected from a set of $N$ randomly sampled crops. Note that $N$ and $k$ are hyperparameters that require tuning. 
In contrast to WCA~\cite{wca}, we set 
$v_j$ in \cref{eq:v_j} to 1, since it represents the self-similarity of each CDA. Finally, the resulting learned alignment score is defined in \cref{eq:wca_sim_func_LTIA}.
\begin{equation}
\psi_{\text{\ours}}\left(x, y | \mathbf{p}, E_v, \textbf{Z}^{*}, \mathcal{C} \right) = \sum_{i=1}^{N} \widetilde{w_i} \widetilde{\Theta}_{ij}|_{j=\mathcal{C}(y)}\mathbb{I}_{\{i \in \mathcal{I}_k\}}\label{eq:wca_sim_func_LTIA}
\end{equation}
where $\mathbb{I}$ denotes the indicator function, and $\mathcal{C}: \mathcal{Y} \to {1 \cdots c}$. 
During training, we leverage the pseudo-label $\hat{y}$ obtained from~\cref{eq:wca_sim_func_LTIA}, as self-supervision for the strongly-augmented counterpart \(\mathcal{A}(x)\), as follows: 
\begin{equation}
    \label{eq:self-training}
    \mathcal{L}_{\text{st}} = - \; \mathbb{E}_{x \in \mathcal{X}_t} \; \left[ \sum_{j=1}^c  \!\mathbb{I}\{\hat{y}=j \!\}\! \; log \left(p_{\mathcal{A}(x)}\right) \right],
\end{equation}
where \(p_{\mathcal{A}(x)} = E_v(\mathcal{A}(x)) \cdot {\textbf{Z}^{*}}^T\) represents the probabilistic output for the strongly-augmented image \(\mathcal{A}(x)\). To mitigate the confirmation bias induced by CLIP~\cite{wang2022debiased}, we adopt the fairness regularization loss $\mathcal{L}_{\text{reg}}$, as proposed by~\cite{li2022masked}: \(\mathcal{L}_{reg} = -\frac{1}{c} \sum_{j=1}^c \log\left(\bar{p}_{\mathcal{A}(x_j)}\right)\)   
where \(\bar{p}_{\mathcal{A}(x)}\) represents a moving average of the model's predictions from the strongly-augmented images across the batch.
This fairness regularization encourages uniform predictions across classes, reducing overfitting to biased pseudo-labels and encouraging balanced adaptation. During inference, we simply compare the query image features with CDA.

\begin{figure}[ht!]
    \centering
    \begin{subfigure}[b]{0.225\textwidth} 
        \centering
        \includegraphics[width=\textwidth]{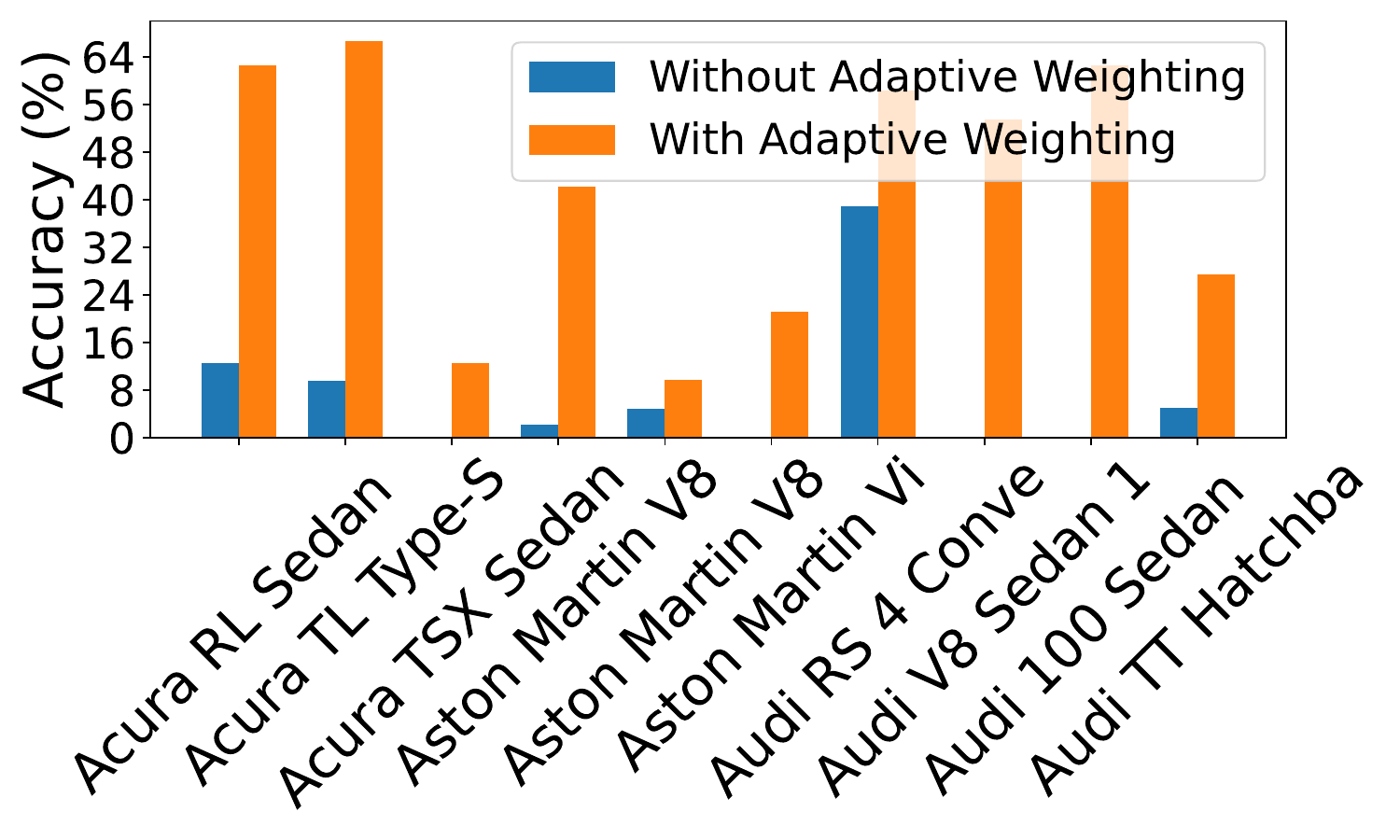}
        \caption{}
        \label{fig:cars_per_acc}
    \end{subfigure}
    \begin{subfigure}[b]{0.225\textwidth}
        \centering
        \includegraphics[width=\textwidth]{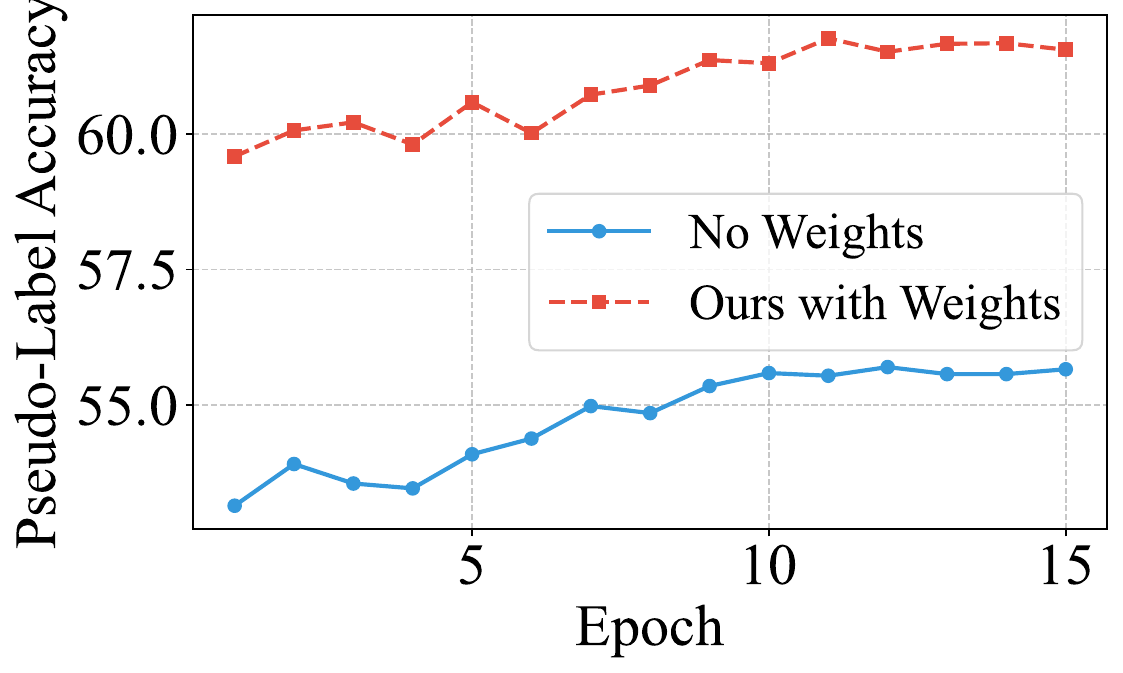}
        \caption{}
        \label{fig:cars_pl_acc}
    \end{subfigure}
    \vspace{-2mm}
    \caption{Visualization of performance improvements from our adaptive weighting mechanism: (a) per-class accuracy and (b) pseudo-label accuracy on the StanfordCars dataset.}
    \label{fig:per_acc_class}
\end{figure}

\noindent 
\textbf{Adaptive Weighting to Mitigate Pseudo-label Noise:}
\label{subsec:Weight}
While \ours shows notable improvements over existing self-training methods, we observe performance degradation on certain fine-grained datasets where the learned descriptions become less distinctive. This results in increased confusion between visually similar classes, sometimes leading to complete misclassifications, as illustrated in~\cref{fig:per_acc_class}. For example, in the StanfordCars dataset, the class `Sedan 2012' is frequently confused with `Sedan 2008', reflecting the difficulty of distinguishing subtle visual differences.

To mitigate this, we introduce a confidence-adjusted weighting mechanism for the classification loss, which adjusts the weight of the pseudo-label $\hat{y}$ based on the top two alignment scores from our LAS $\psi_{\text{\ours}}\left(x, y \mid \mathbf{p}, E_v, \textbf{Z}^{*}, \mathcal{C} \right)$ (see~\cref{fig:main_architecture}: \textcolor[HTML]{82B366}{\dbox{adaptive weighting}}). Specifically, let $S_{i,x}$ and $S_{j,x}$ be the highest and second-highest alignment scores for image $x$ with respect to the CDA representations. The final weight is computed as:
\begin{equation}
    \label{eq:reweighting}
    \gamma_{\text{x}} = S_{i,x} \cdot |S_{i,x} - S_{j,x}|
\end{equation}
where $i$ and $j$ represent the indices of the CDA with the highest and second-highest scores to image $x$, respectively. 
This formulation down-weights uncertain pseudo-labels by penalizing cases where the two top predictions are closely aligned, an indicator of class ambiguity. Consequently, it enhances the model’s robustness to fine-grained distinctions, such as those found in the StanfordCars dataset. 
The self-training loss $\mathcal{L}_{st}$ in \cref{eq:self-training} is reformulated in \cref{eq:weight_self-training}. 
\begin{equation}
    \label{eq:weight_self-training}
    \mathcal{L}_{\text{st}} = - \; \mathbb{E}_{x \in \mathcal{X}_t} \; \left[ \gamma_{x} \cdot \sum_{j=1}^c  \!\mathbb{I}\{\hat{y}=j \!\}\! \; log \left(p_{\mathcal{A}(x)}\right) \right],
    \end{equation}
where $\gamma_{x}$ represents the weight calculated from~\cref{eq:reweighting}. The overall loss function is \(\mathcal{L} = \mathcal{L}_{st} + \mathcal{L}_{reg}\). We provide a detailed pseudo-algorithm of \ours in suppl. materials.

\section{Experiments}
\label{sec:experiments}

\noindent\textbf{Datasets and Baselines:} We extensively evaluate on 13 diverse datasets: 
Caltech101~\cite{fei2004learning},
DTD~\cite{cimpoi2014describing}, 
EuroSAT~\cite{helber2019eurosat}, 
Food101~\cite{bossard2014food}, 
Flowers102~\cite{nilsback2008automated}, 
OxfordPets~\cite{parkhi2012cats}, 
SUN397~\cite{xiao2010sun}, 
StandfordCars~\cite{krause20133d}, 
CIFAR100~\cite{krizhevsky2009learning}, 
UCF101~\cite{soomro2012ucf101},
CUB-200-2011~\cite{cub}, 
AID~\cite{Xia2016AIDAB} 
and RESISC45~\cite{Cheng2017RemoteSI}.
We conduct a comparative analysis against SOTA methods, including three zero-shot approaches: CLIP~\cite{clip}, CuPL~\cite{CuPL}, and WCA~\cite{wca}, alongside five unsupervised adaptation techniques: UPL~\cite{upl}, POUF~\cite{pouf}, LaFTer~\cite{lafter}, ReCLIP~\cite{reclip}, and DPA~\cite{DPA}. 

\begin{figure}[!ht]
    \centering
    \includegraphics[width=0.75\columnwidth]{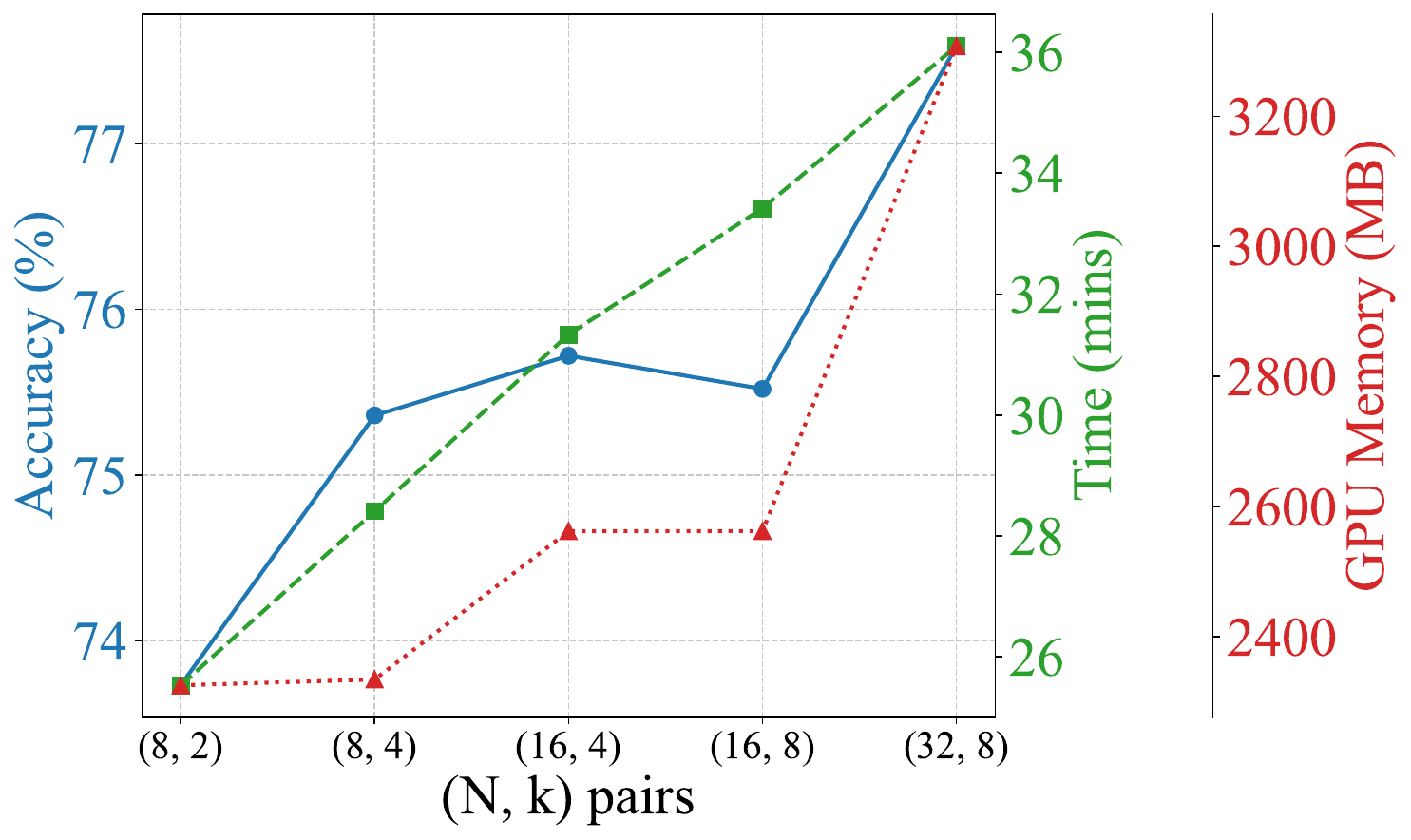}
    \caption{Analysis of accuracy, training time, and GPU memory usage as a function of the number of sampled crops ($N$) and the number of selected top-weighted crops ($k$), evaluated on the Flowers dataset.}
    \label{fig:crop_analysis}
\end{figure}

\noindent \textbf{Implementation Details:} 
We use CLIP~\cite{clip} with a ViT-B/32 backbone for all experiments. In unsupervised fine-tuning, we focus on fine-tuning the layer-normalization weights of the image encoder, a stable and effective strategy for models trained with noisy supervision~\cite{ba2016layer, wang2020tent} along with the CDA. We apply RandomResizedCrop, Flip, and RandAugment~\cite{cubuk2020randaugment} as strong augmentations. The text descriptions used to initialize CDA are derived from CuPL~\cite{CuPL}.
For all the crop-based experiments, we use $(N,k)=(16,4)$, according to the analysis illustrated in~\cref{fig:crop_analysis}. For all experiments, we set the learning rate to $10^{-4}$ for all datasets, except for Food101 and SUN397, where it is set to $10^{-6}$.
For additional implementation details, see suppl. materials.

\noindent\textbf{Overall Results:} 
Results for our method, \ours, are presented in~\cref{table:sota_vit_b_32}. \ours consistently outperforms zero-shot methods—CLIP, CuPL, and WC—by average margins of 11.45\%, 9.89\%, and 10.96\%, respectively, across all datasets. It also surpasses all unsupervised adaptation methods across the 13 datasets. In particular, it outperforms UPL, which optimizes prompts, by 11.69\%. Notably, UPL underperforms even compared to zero-shot CLIP, likely due to its offline pseudo-label generation using CLIP, which may reduce confidence in fine-grained classes~\cite{upl, lafter}.
Furthermore, \ours outperforms POUF, LaFTer, ReCLIP, and DPA with average gains of 9.54\%, 9.82\%, 5.67\%, and 2.78\%, respectively.
Performance comparisons using a different backbone, ViT-B/16, are provided in~\cref{table:sota_vit_b_16}.

\begin{table}[H]
  \centering
  \resizebox{\columnwidth}{!}{
    \begin{tabular}{@{}lccccccc@{}}
      \toprule
      \textbf{Method} & \textbf{DTD} & \textbf{ESAT} & \textbf{Flowers} & \textbf{OxPets} & \textbf{StCars} & \textbf{UCF} & \textbf{Avg} \\
      \midrule
      \rowcolor[gray]{0.9} \multicolumn{8}{c}{\textbf{Zero-shot Methods}} \\
      \midrule
      CLIP~\cite{clip}          & 44.70 & 49.00 & 70.89 & 89.00 & 64.70 & 69.10 & 64.56 \\
      CuPL~\cite{CuPL}           & 54.36 & 58.68 & 73.93 & 91.20 & {66.15} & 69.89 & 69.04 \\
      \midrule
      \rowcolor[gray]{0.9} \multicolumn{8}{c}{\textbf{UA Methods}} \\
      \midrule
      UPL~\cite{upl}            & 45.90 & 55.36 & 73.93 & 87.98 & 60.33 & 67.43 & 65.16 \\
      POUF~\cite{pouf}           & 48.60 & 59.50 & 72.10 & {91.80} & 63.50 & {71.50} & 67.83 \\
      LaFTer~\cite{lafter}         & \underline{54.79} & {72.10} & {75.15} & 85.28 & \underline{64.72} & 67.20 & {69.87} \\
      DPA~\cite{DPA}         & 50.32&	\underline{81.22}	&\underline{78.64}	&\underline{93.35}&	63.97&	\underline{74.44}&	73.66 \\
      \rowcolor[HTML]{FFD6E7} \textbf{\ours} (ours) & \textbf{60.74}	& \textbf{91.74}	&\textbf{80.80}&	\textbf{93.40}&	\textbf{66.16}	&\textbf{79.86}&	\textbf{78.78}\\  
      \bottomrule
    \end{tabular}
  }
  \caption{Top-1 accuracy (\%) comparison of \ours with state-of-the-art unsupervised adaptation methods using CLIP-ViT-B/16.}
  \label{table:sota_vit_b_16}
\end{table}

\noindent\textbf{Analysis of Model Components:} 
We conduct a thorough ablation study to evaluate the contribution of each component in our approach, as shown in~\cref{tab:ablation}.
\noindent An ablation study on \ours's key components demonstrates that removing the Learned Alignment Score (LAS) from \ours causes a substantial performance decrease of 7.04\%, while eliminating only the weighting mechanism reduces performance by 1.84\%. This highlights the critical role of both components, particularly LAS, in effectively generating pseudo-labels from local view features.
Next, we introduce a simplified baseline, \ours-g, which utilizes global visual features for pseudo-labeling instead of local view features. Incorporating LAS with global features results in decreased performance, indicating that LAS is more effective with local features. Incorporating the weighting mechanism alongside LAS yields a 4.19\% accuracy gain, highlighting the crucial role of our weighting mechanism in enhancing model performance.
\noindent Finally, we ablate the weighting mechanism (\ours w/o PL weight) and apply WCA with limited crops for pseudo-label generation. This direct application results in a substantial performance degradation (4.19\%), demonstrating the suboptimal nature of naïve WCA utilization under crop constraints in UA scenarios.
We then enhance the process through three modifications:
(1) selecting the top-k crops, which improves performance by 0.32\%;
(2) incorporating a [CLS]-based crop weighting strategy, increasing performance by 0.07\%; and
(3) integrating LAS via CDA interaction, achieving 71.94\% accuracy.
Combining all three components boosts performance to 74.10\% (\ours w/o PL weight), confirming the effectiveness of our structured refinement pipeline.
For results using different pseudo-labeling strategies, refer to the suppl. materials.

\begin{table}[!ht]
  \centering
  \resizebox{\columnwidth}{!}{
    \begin{tabular}{@{}lccccccc@{}}
      \toprule
      \textbf{Component} & \textbf{DTD} & \textbf{ESAT} & \textbf{Flowers} & \textbf{OxPets} & \textbf{StCars} & \textbf{UCF} & \textbf{Avg} \\
      \midrule
      CLIP~\cite{clip} & 44.63 & 48.90 & 66.42 & 87.46 & 58.70 & 61.86 & 61.33 \\
         \midrule
      \rowcolor[gray]{0.9} \multicolumn{8}{c}{\textbf{Ablations on \ours}} \\
      \midrule
      \textbf{\ours} w/o LAS & 54.68 & 66.96 & 72.27 & 89.56 & 60.23 & 69.63 & 68.89 \\
      \textbf{\ours} w/o PL weight & 59.68 & \underline{90.96} & \underline{75.56} & \textbf{90.92} & 55.84 & 71.61 & \underline{74.10} \\
      \midrule
      \rowcolor[gray]{0.9} \multicolumn{8}{c}{\textbf{Ablations on Global Visual Features for PL}} \\
      \midrule
      \textbf{\ours-g}  & 53.51 &	69.06 & 72.43 &	89.83 & 62.49 & 69.20 & 69.42 \\
      \hspace{0.3cm} + LAS & 57.23 & 72.54 & 72.80 & 89.62 & 55.74 & 68.00 & 69.32 \\
      \hspace{0.3cm} + PL weight & 58.35 & 89.28 & 73.08 & 89.75 & 58.75 & 71.85 & 73.51 \\
      \midrule
       \rowcolor[gray]{0.9} \multicolumn{8}{c}{\textbf{Ablations on \ours (w/o PL weight) PL}} \\
      \midrule
      \textbf{WCA}~\cite{wca} as PL ($\bigstar$) & 54.89 & 65.08 & 72.31 & 89.51 & \underline{63.93} & 69.36 & 69.18 \\
      ($\bigstar$) + Crop Selection & 54.68 &	66.38 &	72.63 & 89.64 & \textbf{63.98} &	69.68 &	69.50 \\
      ($\bigstar$) + \texttt{[CLS]}-based crop weights & 54.95 & 65.48 & 72.68 & 89.51 & 63.31 & 69.60 & 69.26 \\
      ($\bigstar$) + Interaction w/ CDA (LAS) & \underline{61.54} & 78.34 & 74.42	& \underline{90.54} & 54.92 & \underline{71.90} & 71.94 \\        
        \midrule
      \rowcolor[HTML]{FFD6E7} \textbf{\ours} (ours) & \textbf{62.07} & \textbf{91.92} & \textbf{75.72} & 90.52 & 61.83 & \textbf{73.54} & \textbf{75.93}  \\
      \bottomrule
    \end{tabular}
  }
  \caption{Ablation study on model components. \ours-g is a simplified baseline that replaces crop features with the global feature of a weakly augmented image during pseudo-labeling.}
  \label{tab:ablation}
\end{table}

\noindent\textbf{Comparison in Transductive Setting:}
We present a comparative analysis between \ours and SOTA transductive UA methods for VLMs, specifically ReCLIP~\cite{reclip} and DPA~\cite{DPA}, as shown in~\cref{table:reclip}.
Following ReCLIP’s protocol, we adapt CLIP on the unlabeled test split and evaluate on the same split. To ensure a fair comparison, we employ LLM-generated descriptions for training both DPA and ReCLIP, denoted as DPA\textsuperscript{\ddag} and ReCLIP\textsuperscript{\ddag} in~\cref{table:reclip}—thereby avoiding handcrafted prompts.
We also report the zero-shot performance of TransCLIP~\cite{transclip}, referred to as TransCLIP-ZS, which serves as a training-free baseline in the transductive setting. Despite being entirely zero-shot, TransCLIP-ZS underperforms compared to adaptation-based methods, emphasizing the importance of adaptation.
Our method, \ours, consistently outperforms SOTA approaches in both inductive and transductive settings. In the transductive setting, \ours achieves an average accuracy of $75.57\%$, surpassing DPA ($72.86\%$), ReCLIP ($69.27\%$), and the zero-shot TransCLIP-ZS ($66.39\%$), demonstrating superior robustness and adaptability. 

\begin{table}[!ht]
  \centering
  \resizebox{\columnwidth}{!}{
    \begin{tabular}{@{}lcccccccc@{}}
      \toprule
      \textbf{Method} & \textbf{DTD} & \textbf{ESAT} & \textbf{Flowers} & \textbf{OxPets} & \textbf{StCars} & \textbf{UCF} & \textbf{Avg} \\
      \midrule
      \rowcolor[gray]{0.9} \multicolumn{8}{c}{\textbf{Inductive}} \\
      \midrule
      \textbf{ReCLIP}\textsuperscript{\dag}~\cite{reclip} & 53.88 & 70.80 & 72.63 & 87.49 & 59.22 & 67.01 & 68.51 \\
      \textbf{ReCLIP}\textsuperscript{$\ddagger$} & 55.64 & 70.64 &  \underline{75.88} & \underline{90.35} & \underline{60.53} & 70.68 & 70.62 \\
      \textbf{DPA} & 55.96 & \underline{80.04} & 75.56 & 90.71 & 56.83 & 66.69 & 70.85\\      
      \textbf{DPA}\textsuperscript{$\ddagger$} & \underline{58.03} & 77.80 & \textbf{78.97} & 89.94 & 57.26 & \underline{72.75} & \underline{72.46} \\ 
      \rowcolor[HTML]{FFD6E7} \textbf{\ours} (ours) &  \textbf{62.07}	& \textbf{91.92}	&75.72	& \textbf{90.52}	& \textbf{61.83}	& \textbf{73.54}	& \textbf{75.93}      \\
      \midrule
      \rowcolor[gray]{0.9} \multicolumn{8}{c}{\textbf{Transductive}} \\
      \midrule
      \textbf{TransCLIP-ZS}~\cite{transclip} & 51.06 & 53.66 & 74.38 & 89.04 &	\underline{61.15} &	69.07 &	66.39 \\
      \textbf{ReCLIP}~\cite{reclip} & 52.50 & 59.30 & 70.65 & 88.42 & 59.06 & 69.13 & 66.51 \\
      \textbf{ReCLIP}\textsuperscript{$\ddagger$} & 53.19	&67.48	& \underline{74.99}	&90.24&	{59.43}&	{70.26}&	69.27 \\
      \textbf{DPA} & \underline{58.09}	&79.74 &	74.14	&\underline{91.06}&	57.72	&68.73&	71.58 \\ 
      \textbf{DPA}\textsuperscript{$\ddagger$} & 54.73 & \underline{83.13} & \textbf{77.91} & 90.38 & 59.32 & \underline{71.69} & \underline{72.86} \\ 
      \rowcolor[HTML]{FFD6E7} \textbf{\ours} (ours) & \textbf{59.41}	&  \textbf{89.52}	&74.62	&\textbf{91.82	}&\textbf{62.65}	&\textbf{75.42}	&  \textbf{75.57} \\
      \bottomrule
    \end{tabular}
  }
\caption{
Classification accuracy (\%) of \ours, ReCLIP, and DPA in inductive and transductive settings.
\textsuperscript{\dag} indicates that the model is originally designed for transductive training, but we train it in the inductive setting.
\textsuperscript{\ddag} indicates that the model uses handcrafted prompts by default, but we use LLM-generated prompts to ensure a fair comparison with \ours.
We also report the performance of the zero-shot baseline, TransCLIP-ZS, for completeness.
}
  \label{table:reclip}
\end{table}

\noindent\textbf{Analysis of Weighting Components:}
We conduct a comprehensive ablation analysis to evaluate the individual and collective contributions of the weight components $\gamma_{x}$ in~\cref{table:ablation_weight}. 
This systematic investigation involves sequentially removing each component of $\gamma_{x}$ and measuring the resultant impact on \ours's performance.

\begin{table}[!ht]
\centering
  \small
  \setlength{\tabcolsep}{5pt}
  \scalebox{0.85}{
\begin{tabular}{ccccc}
\toprule
\textbf{$S_{i,x}$} & \textbf{$|S_{i,x} - S_{j,x}|$} & \textbf{Avg. Acc. (\%)} \\ 
\midrule
{$\times$} & {$\times$} & 74.10 \\
\checkmark & {$\times$} & 74.63 \\
\rowcolor[HTML]{FFD6E7} \checkmark & \checkmark & \textbf{75.93} \\
\bottomrule
\end{tabular}
}
\caption{Ablation study on the pseudo-label weighting components. Accuracy is averaged across the six ablation datasets.}
\label{table:ablation_weight}
\end{table}

\begin{figure}[!ht]
    \centering
    \includegraphics[width=0.75\columnwidth]{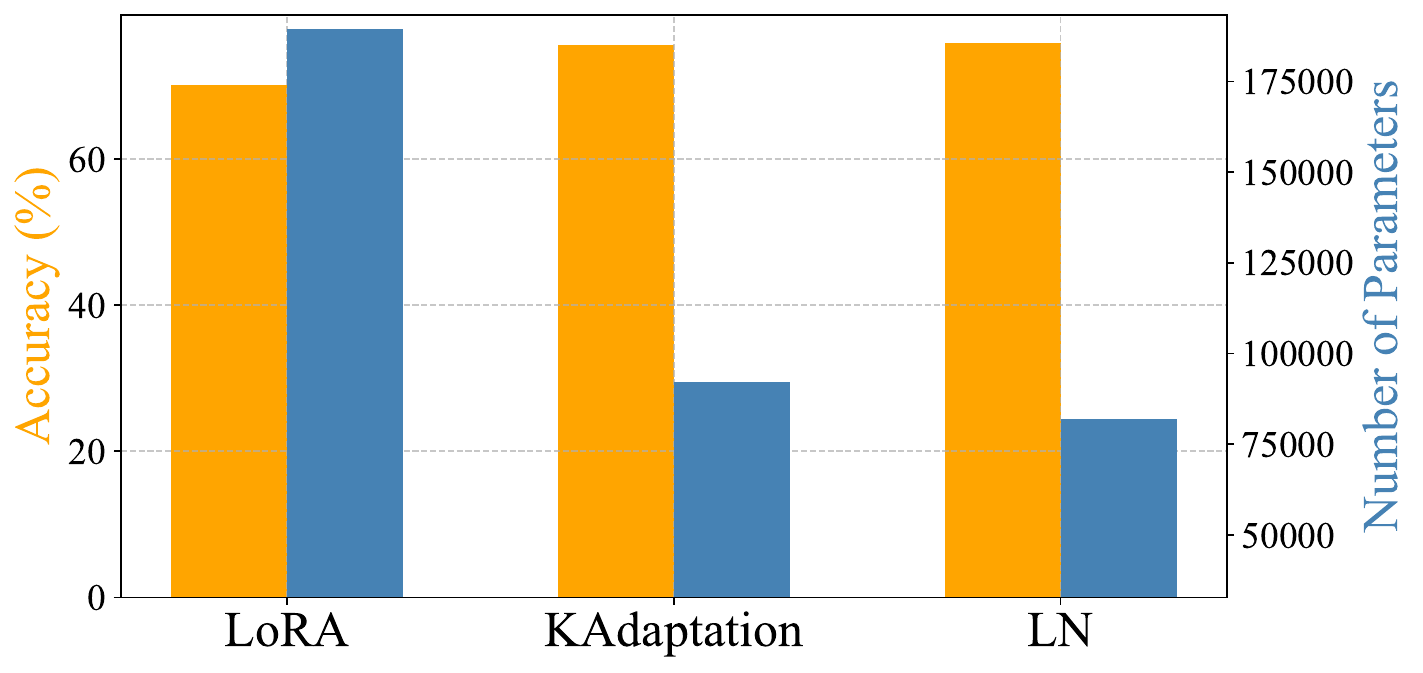}
    \caption{Comparison of accuracy ($\%$) and number of parameters between \ours and different parameter-efficient fine-tuning methods. Accuracy is averaged across the six ablation datasets.}   
    \label{fig:full_tune}
\end{figure}

\noindent\textbf{Evaluation of Fine-Tuning Techniques:}
~\cref{fig:full_tune} compares various CLIP fine-tuning techniques: LoRA\cite{hu2022lora}, K-Adaptation~\cite{he2023parameter}, and our approach (\ours), which targets only the layer normalization weights.
~\cref{fig:full_tune} shows that \ours achieves stable and efficient fine-tuning, striking a balance between accuracy and parameter count.

\noindent\textbf{Hyperparameter Sensitivity Analysis:}
~\cref{fig:crop_analysis} presents a comprehensive sensitivity analysis of two hyperparameters in \ours: the number of sampled crops ($N$) and the number of top-weighted crops ($k$). We evaluate their impact on model accuracy, training efficiency, and GPU memory utilization using the Flowers dataset.

\section{Conclusion}
\label{sec:conclusion}
In this paper, we introduce \ours (Fine-grained Alignment and Interaction Refinement), a novel framework that addresses the challenge of fine-grained classification in fully unlabeled settings for CLIP. \ours pushes the state-of-the-art forward by leveraging cross-modal fine-grained interactions during pseudo-labeling, enabling class-discriminative features that significantly boost CLIP performance. We incorporate an innovative pseudo-label weighting mechanism based on similarity criteria, further refining the alignment between visual and textual features. This refinement enhances the accuracy of the pseudo-labeling process. Our comprehensive evaluation demonstrates that \ours achieves remarkable accuracy improvements across a diverse range of datasets, highlighting its potential to transform the adaptability of CLIP for fine-grained unsupervised adaptation.

\clearpage
\setcounter{page}{1}
\maketitlesupplementary

\noindent
This supplementary material complements the main paper by providing additional insights and details.
\cref{sec:implementation,sec:dataset_statistics} provide implementation details and dataset statistics to support reproducibility.
\cref{sec:zero_shot,sec:vlms,sec:time_complex} discuss extended experimental results, including analyses of zero-shot techniques in pseudo-labeling, comparisons with other VLMs, and an evaluation of model complexity.
Lastly,~\cref{sec:limitations} outlines the limitations of our approach and potential directions for future research. We provide detailed pseudo-code of our method in~\cref{alg:FAIR_train}.

\section{Additional Implementation Details}
\subsection{Implementation and Computation Details:}
\label{sec:implementation}

In our unsupervised fine-tuning approach, we focus on adjusting the layer normalization weights of the image encoder and CDA. This method has been shown to be both effective and stable for adapting models under noisy supervision~\cite{ba2016layer, wang2020tent}. Input images are standardized to a size of 224 × 224. During training, we use RandomResizedCrop, Flip, and RandAugment~\cite{cubuk2020randaugment} as strong augmentation methods, and CenterCrop as the only weak augmentation for \ours, since we augment the image features for pseudo-labeling using RandomCrop, and RandomResizedCrop for \ours-g. We utilize the AdamW optimizer~\cite{loshchilov2017decoupled} with a cosine learning rate schedule. For all experiments, we set the learning rate to $1 \times 10^{-4}$ for all datasets, except for Food101 and SUN397, where it is set to $1 \times 10^{-6}$.
We use a batch size of 32 for all datasets, training for 15 epochs. For all crop-based experiments, we set the hyperparameters to $\alpha = 0.5$ and $\beta = 0.9$, as used in WCA~\cite{wca}. Additionally, we use $(N, k) = (16, 4)$, where $N$ denotes the number of crops generated per image, and $k$ represents the top-$k$ crops selected from a set of $N$ randomly sampled crops.
Our method is implemented using PyTorch, and all experiments are conducted on a single NVIDIA A100-SXM4-40GB GPU. The LLM descriptions used in our study are derived from CuPL~\cite{CuPL}, which automates description generation using carefully designed prompts for LLMs. To ensure fair comparisons, we reproduce the results of SOTA methods using their official codebases. We use VISSL~\cite{goyal2021vissl} to standardize dataset splits across all SOTA methods, ensuring consistency, as different methods use varying dataset partitions.
In both the main paper’s ablation experiments and the supplementary experiments, we evaluate six of the thirteen datasets, excluding Caltech101, AID, CUB, RESIS45, CIFAR-100, Food101, and SUN397. Following ReCLIP’s procedure~\cite{reclip}, we select six diverse datasets—EuroSAT (satellite imagery), UCF-101 (action recognition), Flowers and Cars (fine-grained recognition), DTD (texture), and Pets (intra-class variation)—to balance variety and experimental depth. These smaller datasets enable more extensive experiments while covering diverse domains and difficulty levels for robust generalization evaluation.
A comparison of \ours using RN50 is not feasible, as both \ours and the relevant baselines (e.g., LaFTer~\cite{lafter}, ReCLIP~\cite{reclip}, DPA~\cite{DPA}) are specifically designed for transformer-based image encoders.

\subsection{Dataset Statistics and Splits:}
\label{sec:dataset_statistics}
We conduct experiments on 13 diverse datasets, summarized in~\cref{table:detail}, which outline key details such as the number of text descriptions per class, the number of classes, and the sizes of the training and test sets.

\begin{table}[!ht]
  \centering
  \small
  \begin{tabular}{@{}l@{\hskip 4pt}c@{\hskip 4pt}c@{\hskip 4pt}c@{\hskip 4pt}c@{\hskip 4pt}c@{}}
    \toprule
    \textbf{Dataset} & \textbf{Abbr.} & \textbf{Desc/Class} & \textbf{Classes} & \textbf{Train} & \textbf{Test} \\
    \midrule
    Caltech101 & Caltech & 30 & 100 & 4,403 & 6,645 \\
    DTD & DTD & 60 & 47 & 3,760 & 1,880 \\
    EuroSAT & ESAT & 25 & 10 & 10,000 & 5,000 \\
    Food101 & Food & 30 & 101 & 75,750 & 25,250 \\
    Flowers102 & Flower & 20 & 102 & 4,093 & 2,463 \\
    Oxford Pets & OxPets & 20 & 37 & 3,680 & 3,669 \\
    SUN397 & SUN & 30 & 397 & 76,129 & 21,758 \\
    Stanford Cars & StCars & 90 & 196 & 8,144 & 8,041 \\
    CIFAR10 & CIFAR10 & 30 & 10 & 50,000 & 10,000 \\
    CIFAR100 & CIFAR100 & 40 & 100 & 50,000 & 10,000 \\
    UCF101 & UCF & 50 & 101 & 9,537 & 3,783 \\
    CUB-200-2011 & CUB & 31-40 & 200 & 5,994 & 5,794 \\
    RESISC45 & UCF & 50 & 45 & 25,200 & 6,300 \\
    AID & AID & 50 & 30 & 7,000 & 3,000 \\
    \bottomrule
  \end{tabular}
  \caption{Detailed dataset statistics.}
  \label{table:detail}
\end{table}

\section{Additional Experiments and Comparisons}

\subsection{Empirical Study of Zero-Shot Techniques in Pseudo-labeling}
\label{sec:zero_shot}
We conduct experiments with CODER~\cite{encoder}, a recent method that has demonstrated SOTA performance in zero-shot learning, to further evaluate the effectiveness of these techniques for pseudo-labeling, particularly focusing on the benefits of fine-grained interactions achieved through localized image features and self-learned class description anchors in our approach. CODER employs an LLM to generate five distinct types of prompts, which are then used to construct dense, class-specific textual representations. These descriptions are systematically compared to global image features, enabling a unimodal fine-grained interaction. CODER’s standout contribution lies in using an Auto Text Generator (ATG) to effectively utilize these rich textual descriptions. 
Compared to CuPL~\cite{CuPL}, CODER utilizes denser and more systematic textual descriptions for each class. This detailed representation provides a performance edge in tasks requiring fine-grained image classification. For example, CODER achieves an average improvement of 1.38\% over \ours-g (68.87\% to 70.25\%), as shown in~\cref{tab:coder_comparison}. This improvement highlights the value of dense textual neighbors and their integration with visual features, akin to our \ours-g approach but with better-calibrated textual embeddings.
While the dense textual embeddings improve CODER’s performance as a pseudo-labeler, our proposed method, \ours, achieves SOTA results, surpassing CODER by a significant margin (5.13\% average accuracy). This improvement is attributed to the fine-grained interactions between localized image features and learned CDA during pseudo-labeling. Our method’s robust design enables it to outperform other approaches across multiple benchmarks, demonstrating the efficacy of \ours as a generalized pseudo-labeling framework.

\begin{table}[!ht]
  \centering
  \resizebox{\columnwidth}{!}{
    \begin{tabular}{@{}lccccccc@{}}
      \toprule
      \textbf{Method} & \textbf{DTD} & \textbf{ESAT} & \textbf{Flowers} & \textbf{OxPets} & \textbf{StCars} & \textbf{UCF} & \textbf{Avg} \\
      \midrule
      \ours (w/ WCA PL) & 55.16 & 66.10 & 71.86 & 89.45 & \underline{60.68} & \underline{69.39} & 68.77 \\
      \ours-g & 53.98 & 67.92 & 72.47 & \underline{89.94} & 59.72 & 69.18 & 68.87 \\
      \ours (w/ CODER PL) & \underline{55.80} & \underline{77.30} & \underline{73.69} & 86.56 & 59.33 & 68.81 & \underline{70.25} \\
      \rowcolor[HTML]{FFD6E7} \textbf{\ours} (ours) & \textbf{62.07} & \textbf{91.92} & \textbf{75.72} & \textbf{90.52} & \textbf{61.83} & \textbf{73.54} & \textbf{75.93} \\
      \bottomrule
    \end{tabular}    
  }
  \caption{Comparison of average top-1 accuracy between \ours and other SOTA zero-shot methods used as pseudo-labelers. Notably, \ours-g, our simplified baseline that uses global views instead of localized crops, incorporates CuPL~\cite{CuPL} as its pseudo-labeling method. \ours’s pseudo-labeling weight has been used in all three other methods for a fair comparison. In contrast to the other three methods, \ours refines the alignment score function using fine-grained interactions during pseudo-labeling.}
  \label{tab:coder_comparison}
\end{table}

\subsection{Comparison with latest VLMs}
\label{sec:vlms}
\Cref{tab:metaclip_results} presents a comparison between our method, \ours, and state-of-the-art (SOTA) methods using a different vision-language model (VLM), MetaCLIP~\cite{xu2023demystifying} (ViT-B/32). As shown in \Cref{tab:metaclip_results}, \ours consistently outperforms SOTA methods across the ablation datasets. Notably, \ours maintains its effectiveness even when the underlying VLM changes, demonstrating robustness and adaptability across diverse VLMs.

\begin{figure*}[!ht]
    \centering
    \begin{subfigure}[b]{0.45\textwidth} 
        \centering
        \includegraphics[width=\textwidth]{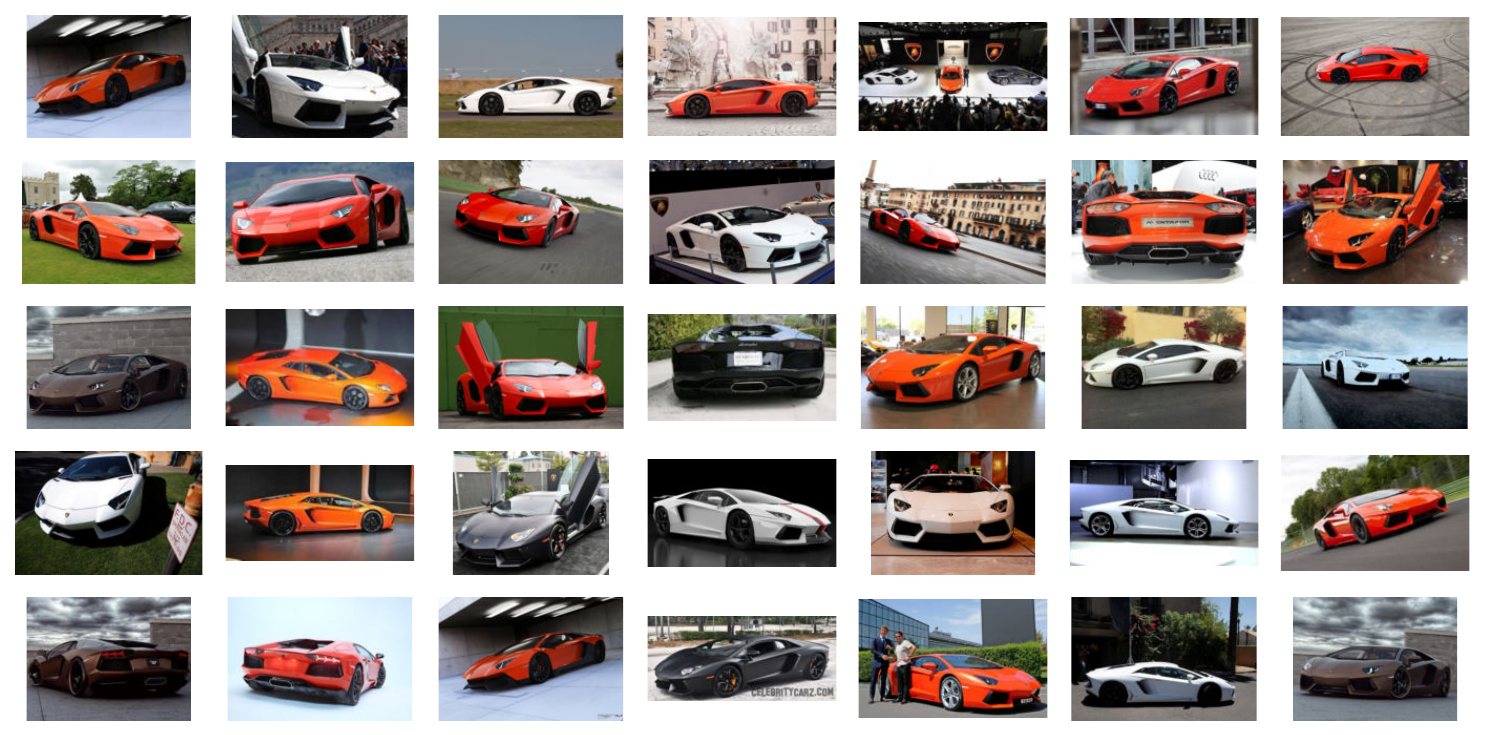}
        \caption{}
        \label{fig:aventador}
    \end{subfigure}
    \begin{subfigure}[b]{0.45\textwidth}
        \centering
        \includegraphics[width=\textwidth]{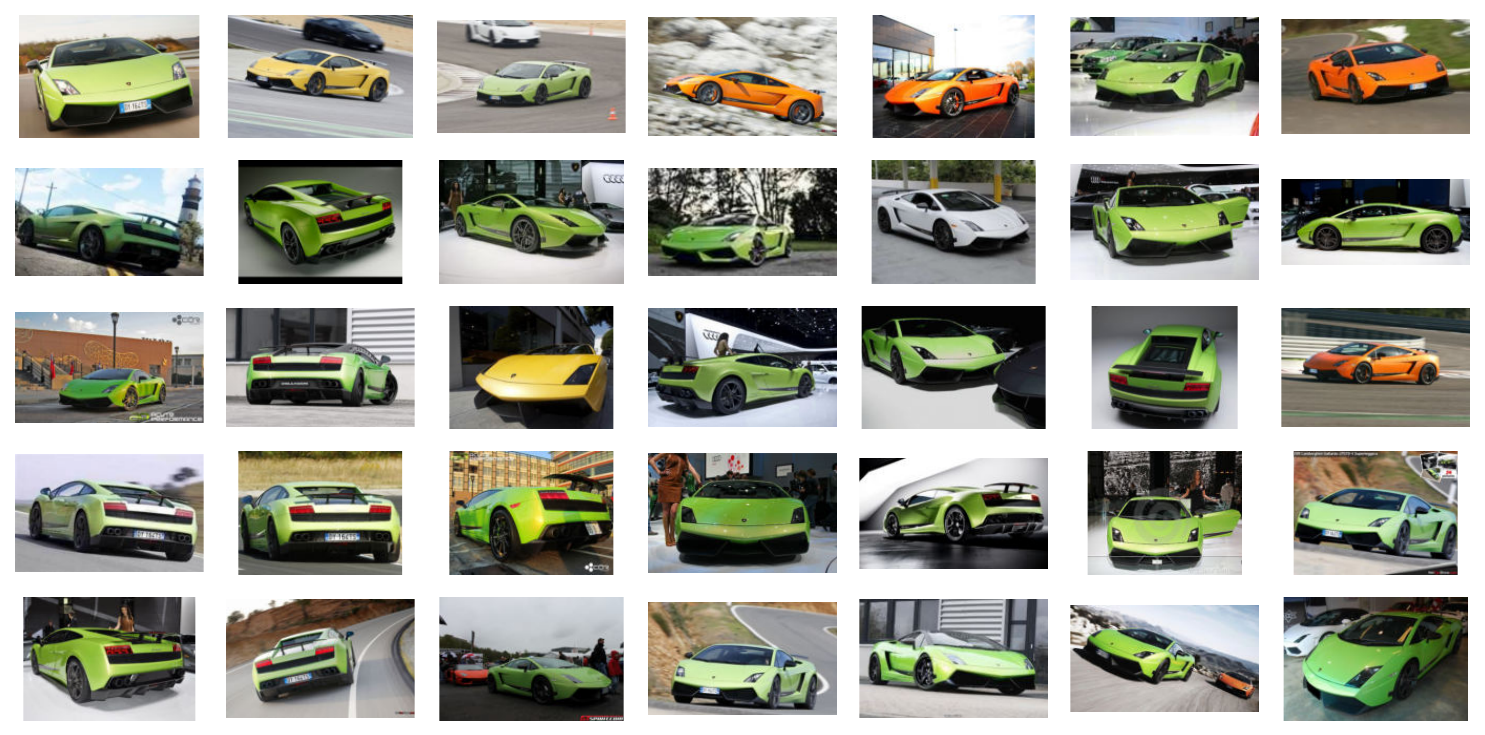}
        \caption{}
        \label{fig:gallardo}
    \end{subfigure}
    \begin{subfigure}[b]{0.45\textwidth}
        \centering
        \begin{tcolorbox}[colback=white!95!black, colframe=black, fontupper=\scriptsize]
        1. This is a photo of a Lamborghini Aventador Coupe from 2012.
        
        2. The Lamborghini Aventador Coupe 2012 is a type of car that looks like a small sports car.
        
        3. The Lamborghini Aventador Coupe 2012 can be identified by its distinctive style and performance.
        \end{tcolorbox}
        \caption{}
        \label{fig:aventador_top3}
    \end{subfigure}
    \begin{subfigure}[b]{0.45\textwidth}
        \centering
        \begin{tcolorbox}[colback=white!95!black, colframe=black, fontupper=\scriptsize]
        1. This is a photo of a Lamborghini Gallardo LP 570-4 Superleggera.
        
        2. The exterior of the Lamborghini Gallardo LP 570-4 Superleggera 2012 is characterized by its aggressive and sporty design.
        
        3. The Lamborghini Gallardo LP 570-4 Superleggera 2012 is a two-seater supercar that is equipped with a 5.
        \end{tcolorbox}
        \caption{}
        \label{fig:gallardo_top3}
    \end{subfigure}
    \caption{Comparison of images between two visually similar car models from the same manufacturer: (a) Lamborghini Aventador Coupe 2012 and (b) Lamborghini Gallardo LP 570-4 Superleggera 2012. These images highlight subtle design differences, posing challenges for accurate classification in fine-grained recognition tasks. (c) and (d) show the top three descriptions that best match a randomly selected image from each class.
    }
\end{figure*}

\begin{figure*}[!ht]
    \centering
    \includegraphics[width=\textwidth]{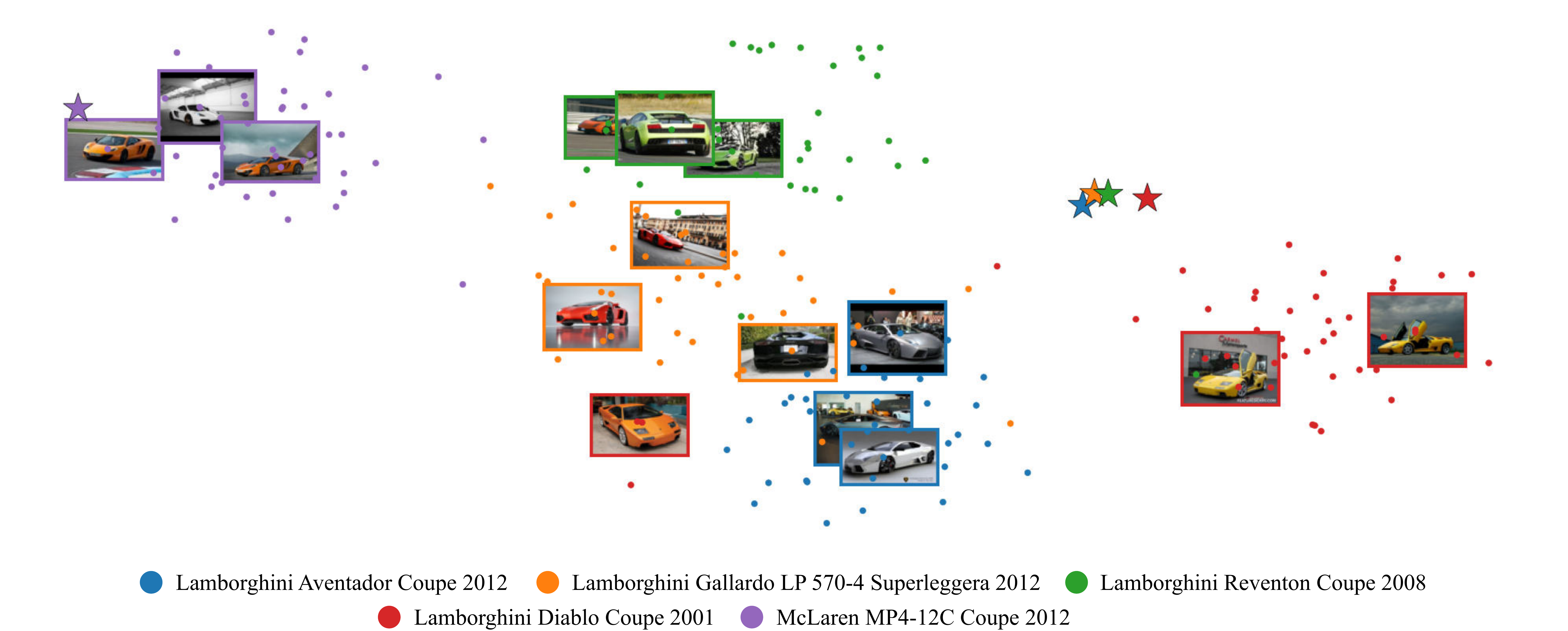}
    \caption{t-SNE visualization of embeddings from five similar car classes in the StanfordCars~\cite{krause20133d} dataset. Dots ($\newmoon$) represent image embeddings, while stars ($\bigstar$) denote Class Description Anchors. Overlaid images correspond to three random points per class, illustrating the overlap and proximity between embeddings of visually similar car models.}
    \label{fig:tsne_cars}
\end{figure*}

\begin{table}[!ht]
\centering
\small
\resizebox{\columnwidth}{!}{
\begin{tabular}{@{}lccccccc@{}}
\toprule
    \textbf{Method} & \textbf{DTD} & \textbf{ESAT} & \textbf{Flowers} & \textbf{Pets} & \textbf{Cars} & \textbf{UCF} & \textbf{Avg} \\
    \midrule
    ZS CuPL & 60.96 & 51.29 & 69.91 & 88.50 & 68.23 & 64.10 & 67.17 \\
    DPA     & 56.90 & 88.14 & \textbf{76.86} & 89.80 & 69.40 & 72.10 & 75.53 \\
    \rowcolor[HTML]{FFD6E7} \textbf{\ours} (ours) & \textbf{70.05} & \textbf{89.72} & 74.38 & \textbf{90.73} & \textbf{70.87} & \textbf{78.01} & \textbf{78.96} \\
\bottomrule
\end{tabular}
}
\caption{Performance comparison of methods using MetaCLIP (ViT-B/32)}
\label{tab:metaclip_results}
\end{table}

\subsection{Comparison of Model Complexity} 
\label{sec:time_complex}

\Cref{table:time} provides a detailed comparison of \ours with SOTA methods in terms of computational cost, measured by the total number of optimized parameters. 
Since \ours backpropagates only through the top-$k$ most informative image patches, \ours enables competitive performance with fewer optimized parameters. Unlike ReCLIP, which requires backpropagation through two encoders and label propagation, \ours trains a single encoder and achieves faster convergence (15 vs. 50 epochs), significantly improving computational efficiency even when accounting for the image cropping overhead. 

Furthermore, \Cref{table:lora} highlights that both fine-tuning strategies—\ours-LN and \ours-KAdaptation—yield nearly identical performance across datasets, confirming the adaptability and robustness of our approach across various parameter-efficient configurations.

\begin{table}[!ht]
  \centering
  \small
  \setlength{\tabcolsep}{4pt}
  \renewcommand{\arraystretch}{1.1}
  \begin{tabular}{@{}lcc@{}}
    \toprule
    \textbf{Method} & \textbf{\#Params} & \textbf{Acc. (\%)} \\
    \midrule
    Zero-shot CLIP       & 0       & 43.84 \\
    UPL                  & 8.2K    & 51.88 \\
    POUF                 & 2.0K    & 62.90 \\
    LaFTer               & 1.14M   & 69.96 \\
    ReCLIP               & 65.5K   & 70.80 \\
    DPA                  & 45.1K   & 79.94 \\
    \rowcolor[HTML]{FFD6E7} \textbf{\ours{} (ours)} & 45.1K   & 91.92 \\
    \bottomrule
  \end{tabular}
  \caption{Accuracy and trainable parameters on EuroSAT (ViT-B/32).}
  \label{table:time}
\end{table}

\begin{table}[!ht]
  \centering
  \resizebox{\columnwidth}{!}{
    \begin{tabular}{@{}lccccccc@{}}
      \toprule
      \textbf{Method} & \textbf{DTD} & \textbf{ESAT} & \textbf{Flower} & \textbf{OxPets} & \textbf{StCars} & \textbf{UCF101} & \textbf{Avg} \\
      \midrule
      \ours-LoRA~\cite{hu2022lora} & 58.30 & 71.82 & 72.51 & 89.78 & 57.41 & 71.27 & 70.18 \\      
      \ours-KAdaptation~\cite{he2023parameter} & 61.86 & 90.62 & \textbf{76.86} & \textbf{90.81} & \underline{60.50} & \underline{72.80} & \underline{75.58} \\
      \rowcolor[HTML]{FFD6E7} \ours-LN (ours)& \textbf{62.07} & \textbf{91.92} & \underline{75.72} & \underline{90.52} & \textbf{61.83} & \textbf{73.54} & \textbf{75.93} \\
      \bottomrule
    \end{tabular}
  }
  \caption{Accuracy (\%) comparison of CLIP fine-tuning techniques on \ours. All experiments use ViT-B/32 as the backbone.}
  \label{table:lora}
\end{table}

\section{Limitations and Future Work}
\label{sec:limitations}

While \ours demonstrates robust performance on fine-grained datasets such as EuroSAT~\cite{helber2019eurosat}, it exhibits limitations on the StanfordCars dataset~\cite{krause20133d}. This dataset presents unique challenges stemming from its detailed intra-class variations (e.g., different car models from the same manufacturer) and high inter-class similarities (e.g., visually analogous cars from different manufacturers). In the absence of labeled data, pseudo-labeling becomes unreliable when differentiating between visually similar classes, potentially causing cascading errors during training. These difficulties are compounded by domain-specific biases in pretrained models, which are typically optimized for more common object categories and thus struggle to capture the nuanced distinctions required in fine-grained recognition tasks like those in the StanfordCars dataset. Consequently, these factors collectively constrain our method's generalizability and performance in highly specialized visual recognition contexts.

An example of the challenges faced in fine-grained classification is the comparison between Lamborghini Aventador Coupe 2012' and Lamborghini Gallardo LP 570-4 Superleggera 2012', as shown in~\cref{fig:aventador} and~\cref{fig:gallardo}, respectively. These Lamborghini models exhibit striking stylistic similarities, with distinctions often limited to subtle design elements like headlight shapes or air intake structures. This difficulty is further illustrated in~\cref{fig:tsne_cars}, where embeddings of these two classes, alongside three other visually similar car models, are projected in a t-SNE space. The figure demonstrates significant overlap in embedding space, with multiple instances of inter-class confusion. This overlap is exacerbated in self-trained settings, as pseudo-labeling struggles to delineate fine-grained distinctions without labeled data. The embedding proximity highlights the inherent challenge of disentangling subtle visual differences when relying on self-supervised pseudo-labels.

To mitigate these limitations, future work should focus on generating denser, fine-grained descriptions that capture subtle class-specific features, such as detailed structural components or stylistic cues, and on improving pseudo-label generation for fine-grained recognition tasks. As shown in~\cref{fig:aventador_top3} and~\cref{fig:gallardo_top3}, current LLM-generated descriptions~\cite{CuPL} lack the necessary specificity to differentiate between visually similar subclasses, thereby exacerbating intra-class ambiguity. Further advancements could involve domain-adaptive fine-tuning to address the domain-specific biases in pretrained models, as well as incorporating attention mechanisms to refine the model’s focus on relevant details.

\begin{algorithm*}[!ht]
    \caption{\ours self-training}\label{alg:FAIR_train}
    \begin{minipage}{\textwidth}
        \begin{algorithmic}[1]
            \Require CLIP vision encoder, $E_v^\Theta$ where $\Theta$ represents all the affine parameters in the LayerNorm layers;
            \Statex \hspace{2.5em} Unlabeled images of a target dataset $\mathcal{X}_t = \{x_i\}_{i=1}^N$;
            \Statex \hspace{2.5em} An LLM model $h(\cdot)$;
            \Statex \hspace{2.5em} Set of class names $\mathcal{Y}$;
            \Statex \hspace{2.5em} Class names to integer map, $\mathcal{C}: \mathcal{Y} \to {1 \cdots C}$;
            \Statex \hspace{2.5em} Weak augmentation $\alpha(\cdot)$;
            \Statex \hspace{2.5em} Strong augmentation $\mathcal{A}(\cdot)$;
            \Statex \hspace{2.5em} Number of epochs $\texttt{MaxEpochs}$;
            \Statex \hspace{2.5em} Batch size $\texttt{B}$
            \Function{InitCDA}{$E_t$, $\mathcal{Y}$, h}
                \State $\mathbf{Z}^{*} \gets \{\emptyset\}_{j=1}^C$
                \ForEach{$y \in \mathcal{Y}$}
                    \State $\mathbf{t} \gets h(y)$ \Comment{Prompt the LLM to extract $M$ number of descriptions for class $y$}
                    \State $\mathbf{Z}_j \gets \frac{1}{M}\sum_{i=1}^M E_t(\mathbf{t})$ \Comment{Take the average of the description embedding $Z^{*}_j \in \mathbb{R}^{1\times d}$ for class $y$}
                \EndFor
                \State \Return $\mathbf{Z}^{*}$
            \EndFunction
            ~\\
            \Function{AdaptiveWeight}{$\psi_{\ours}$}
                \State $S_{i,x}, S_{j,x} \gets \text{top-2}(\psi_{\ours})$ \Comment{Get the top-2 logits by evaluating the \ours similarity function}
                \State $\gamma_x \gets S_{i,x}\cdot(S_{i,x}-S_{j,x})$
                \State \Return $\gamma_x$
            \EndFunction
            ~\\
            \State $\mathbf{Z}^{*} \gets$ \Call{InitCDA}{$E_t$, $\mathcal{Y}$, h} \Comment{Initialize CDA}
            \For{$\texttt{epoch} \gets 1$ to $\texttt{MaxEpochs}$}
                \State $\textbf{x} \gets$ \Call{SampleMiniBatch}{$\mathcal{X}_t$, $B$} \Comment{$\mathbf{x} \in \mathbb{R}^{B\times W\times H\times 3}$}
                ~\\
                \State With \textbf{no Back-Propagation}:
                    \State \hspace{2.5em} $\mathbf{p}(\mathbf{x}) \gets \left\{ p_i = \phi(\alpha(\mathbf{x}), \lambda_i \min(W, H)) \mid i = 1, \dots, N \right\}$ \Comment{Extract $N$ random crops from the weakly augmented image $\alpha(x)$}
                    \State \hspace{2.5em} $\mathbf{f}, \mathbf{f}^{[\texttt{CLS}]} \gets E_v^\Theta\left(p(\mathbf{x})\right)$ \Comment{Local view representation features $\in \mathbb{R}^{B\times d}$}
                    \State \hspace{2.5em} $\widetilde{\Theta} \gets
                    \begin{bmatrix} \text{sim}(\mathbf{f}_1, \textbf{Z}^{*}_1) & \cdots & \text{sim}(\mathbf{f}_1, \textbf{Z}^{*}_C) \\ \vdots & \ddots & \vdots \\ \text{sim}(\mathbf{f}_{N}, \textbf{Z}^{*}_1) & \cdots & \text{sim}(\mathbf{f}_{N}, \textbf{Z}^{*}_C)
                    \end{bmatrix}$
                    \State \hspace{2.5em} $\widetilde{w_i} \gets \frac{\text{sim}(f^{\texttt{[CLS]}}, f_i^{\texttt{[CLS]}})}{\sum_{l=1}^{N}\text{sim}(f^{\texttt{[CLS]}}, f_{l}^{\texttt{[CLS]}})}$
                    \State \hspace{2.5em} $\mathcal{I}_k \gets \text{argsort}(\mathcal{\widetilde{W}})[:k]$ \Comment{Select the top-$k$ crop indices based on $\widetilde{\mathcal{W}}$}
                    \State \hspace{2.5em} $\psi_{\text{\ours}}\left(x, y | \mathbf{p}, E_v, \textbf{Z}^{*}, \mathcal{C} \right) \gets \sum_{i=1}^{N} \widetilde{w_i} \widetilde{\Theta}_{ij}|_{j=\mathcal{C}(y)}\mathbb{I}_{\{i \in \mathcal{I}_k\}}$ \Comment{Evaluate \ours similarity function}
                    \State \hspace{2.5em} $\hat{y} \gets \arg\max_{y \in \mathcal{Y}} (\psi_{\text{\ours}}\left(x, y | \mathbf{p}, E_v, \textbf{Z}^{*}, \mathcal{C} \right)$ \Comment{Compute the pseudo-labels}
                    \State \hspace{2.5em} $\mathbf{\gamma_x} \gets$  \Call{AdaptiveWeight}{$\psi_{\text{\ours}}\left(x, y | \mathbf{p}, E_v, \textbf{Z}^{*}, \mathcal{C} \right)$} \Comment{Compute the adaptive weights}
                ~\\
                \State $p_{\mathcal{A}(x)} \gets \text{softmax}(\text{sim}(E_v^\Theta\left(\mathcal{A}(\mathbf{x})\right), \mathbf{Z}^{*T}), axis=1)$ \Comment{Strongly-augmented counterpart}
                \State $\mathcal{L}_{st} \gets \gamma_x \cdot \text{cross\_entropy}(p_{\mathcal{A}(x)}, \mathbf{\hat{y}})$ \Comment{Self-training loss}
                \State $\mathcal{L}_{reg} \gets -\frac{1}{C} \sum_{j=1}^C \log\left(\bar{p}_{\mathcal{A}(x),j}\right)$ \Comment{Fairness regularization loss}
                \State $\mathcal{L} \gets \mathcal{L}_{st} + \mathcal{L}_{reg}$
                \State \textbf{Back-Propagate} over $\Theta$ and $\mathbf{Z}^{*}$ on $\mathcal{L}$
            \EndFor
        \end{algorithmic}
    \end{minipage}
\end{algorithm*}

{
    \small
    \bibliographystyle{ieee_fullname}
    \bibliography{main}
}

\end{document}